\documentclass[11pt]{article}
\usepackage[margin=1in]{geometry}

\usepackage{amsmath,amssymb}
\usepackage{tikz-cd}
\usepackage{multicol}
\usepackage{url}
\usepackage{array}
\usepackage{multirow}
\usepackage{graphicx}
\usepackage{fancyhdr}
\usetikzlibrary{positioning,arrows.meta,fit}

\fancypagestyle{plain}{%
  \fancyhf{}
  \fancyhead[L]{\includegraphics[height=0.42cm]{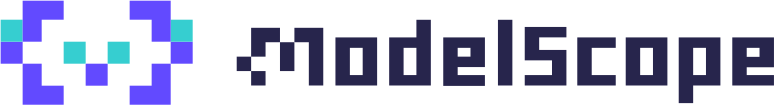}}
  \fancyfoot[C]{\thepage}

}

\title{Diffusion Templates: A Unified Plugin Framework for\\Controllable Diffusion}
\author{Zhongjie Duan$^1$, Hong Zhang and Yingda Chen\\ModelScope Team, Alibaba Group\\$^1$\texttt{duanzhongjie.dzj@alibaba-inc.com}}
\date{}

\begin{document}
\maketitle

\begin{abstract}
\noindent
Controllable diffusion methods have substantially expanded the practical utility of diffusion models, but they are typically developed as isolated, backbone-specific systems with incompatible training pipelines, parameter formats, and runtime hooks. This fragmentation makes it difficult to reuse infrastructure across tasks, transfer capabilities across backbones, or compose multiple controls within a single generation pipeline. We present Diffusion Templates, a unified and open plugin framework that decouples base-model inference from controllable capability injection. The framework is organized around three components: Template models that map arbitrary task-specific inputs to an intermediate capability representation, a Template cache that functions as a standardized interface for capability injection, and a Template pipeline that loads, merges, and injects one or more Template caches into the base diffusion runtime. Because the interface is defined at the systems level rather than tied to a specific control architecture, heterogeneous capability carriers such as KV-Cache and LoRA can be supported under the same abstraction. Based on this design, we build a diverse model zoo spanning structural control, brightness adjustment, color adjustment, image editing, super-resolution, sharpness enhancement, aesthetic alignment, content reference, local inpainting, and age control. These case studies show that Diffusion Templates can unify a broad range of controllable generation tasks while preserving modularity, composability, and practical extensibility across rapidly evolving diffusion backbones. All resources will be open sourced, including code\footnote{\url{https://github.com/modelscope/DiffSynth-Studio}}, models\footnote{\url{https://modelscope.cn/collections/DiffSynth-Studio/KleinBase4B-Templates}}, and datasets\footnote{\url{https://modelscope.cn/collections/DiffSynth-Studio/ImagePulseV2}}.
\end{abstract}

\section{Introduction}

Diffusion models have rapidly become a dominant foundation for visual generation, spanning high-quality image synthesis, image editing, and increasingly video generation~\cite{rombach2022ldm,podell2023sdxl,esser2024sd3,peebles2023dit,blackforest2024flux,wan2025wan}. As these backbones improve, practical applications demand richer forms of control over structure, appearance, identity, editing intent, style, and other task-specific factors. This demand has driven a broad family of controllable generation methods, including parameter-efficient adaptation methods such as LoRA~\cite{hu2022lora}, personalization methods such as Textual Inversion and DreamBooth~\cite{gal2022textualinversion,ruiz2023dreambooth}, and conditional control modules such as ControlNet, T2I-Adapter, and IP-Adapter~\cite{zhang2023controlnet,mou2023t2iadapter,ye2023ipadapter}. These methods are highly effective, but they are usually developed as isolated systems built around particular model architectures, condition types, and training recipes.

This fragmentation creates an increasingly important systems bottleneck for controllable diffusion. In training, different control methods often require different model modifications, parameterizations, preprocessing code, and optimization objectives, which makes it difficult to reuse infrastructure across tasks or transfer a capability from one backbone to another. In deployment, each method exposes its own runtime hooks and parameter formats, so integrating a new control often means editing the pipeline itself rather than simply loading a reusable module. The difficulty becomes more severe when multiple controls must be enabled together: their conditioning pathways may compete for the same internal activations, require incompatible input formats, or depend on ad hoc fusion logic, making conflict resolution and joint scheduling largely a handcrafted engineering problem. As a result, today's controllable diffusion ecosystem remains powerful but poorly modularized.

In this paper, we argue that controllable generation capabilities should be treated as reusable plugins rather than backbone-specific modifications. We present \textbf{Diffusion Templates}, a unified and open plugin framework that decouples base-model inference from controllable capability injection. Our central claim is that many controllable diffusion methods can be reformulated as independent capability modules with a common systems interface, without restricting their model architecture or the format of their control conditions. Under this view, the base diffusion model remains responsible for generation quality, while each controllable capability is packaged as a Template model that can be trained, loaded, and composed independently.

The framework is organized around three components. A Template model takes task-specific input, such as structural signals, scalar attributes, reference images, or other control conditions, and converts it into an intermediate capability representation called a Template cache. This cache is intentionally defined at the interface level rather than tied to a single internal form, so in practice it can be realized through different mediating representations, such as KV-Cache~\cite{kwon2023vllm}, LoRA~\cite{hu2022lora}, or other possible capability states. A Template pipeline then loads one or more Template models, collects their emitted caches, and injects them into the base diffusion runtime during generation. This separation cleanly factorizes controllable generation into two layers: capability construction and capability consumption. Compared with prior controllable diffusion models such as ControlNet or IP-Adapter, the key difference is that Diffusion Templates does not prescribe a specific control architecture or a fixed condition format. Instead, it defines a general interface through which heterogeneous control modules can interact with the same diffusion backbone while preserving strong empirical performance.

Our design is loosely inspired by plugin abstractions in LLM systems, where standardized interfaces have made it possible to extend a strong base model with independently packaged capabilities~\cite{anthropic2024mcp,anthropic2025skills}. We adopt this systems principle for diffusion models, but our focus is not on analogy for its own sake. Rather, the motivation is practical: once controllable capabilities are exposed through a stable interface, they become easier to train, reuse, combine, schedule, and maintain across a rapidly evolving family of diffusion backbones.

Based on this framework, we train and release a diverse model zoo of Template models spanning structural control, brightness adjustment, color adjustment, image editing, super-resolution, sharpness enhancement, aesthetic alignment, content reference, local inpainting, and age control. Together, these case studies cover heterogeneous inputs, lightweight attribute controls, and more demanding image-conditioned tasks under the same runtime abstraction. They show that Diffusion Templates can unify a wide range of controllable generation capabilities without repeatedly redesigning the underlying diffusion pipeline.

In summary, this paper makes the following contributions:
\begin{itemize}
    \item We propose Diffusion Templates, a unified plugin framework for controllable diffusion models that decouples base-model inference from capability injection and provides a common interface for training, loading, and composing heterogeneous control modules.
    \item We train and release a diverse set of Template models spanning structural control, brightness adjustment, color adjustment, image editing, super-resolution, sharpness enhancement, aesthetic alignment, content reference, local inpainting, and age control, demonstrating the generality and practical potential of the framework across varied controllable generation tasks.
\end{itemize}

\section{Related Work}

\subsection{Diffusion Foundation Models}
Diffusion foundation models have rapidly progressed from early denoising formulations (DDPM~\cite{ho2020ddpm}, DDIM~\cite{song2021ddim}) to large-scale latent and transformer-based foundation models. A key milestone is LDM~\cite{rombach2022ldm}, which established the practical latent diffusion paradigm and made high-quality generation computationally feasible at scale. Building on this line of work, the Stable Diffusion family has evolved from early Stable Diffusion releases to SD-XL~\cite{podell2023sdxl} and Stable Diffusion 3~\cite{esser2024sd3}, continuously improving semantic alignment, typography, and high-resolution synthesis quality. At the architectural level, DiT (Diffusion Transformer)~\cite{peebles2023dit} further accelerated the shift toward transformer-native diffusion backbones. In parallel, the open ecosystem has become increasingly diverse, with strong image-generation foundations such as FLUX~\cite{blackforest2024flux}, Hunyuan-Image~\cite{tencent2024hunyuanimage}, PixArt~\cite{chen2024pixart}, SANA~\cite{xie2025sana}, and Qwen-Image~\cite{qwen2025qwenimage}. Video generation is advancing in a similar direction, represented by Wan~\cite{wan2025wan}, LTX~\cite{hacohen2026ltx2}, and Hunyuan-Video~\cite{kong2025hunyuanvideo}, which push diffusion foundations from static image synthesis toward temporally coherent generation.

We aim to expose these powerful base models through reusable control interfaces, so their capabilities can be efficiently transferred to downstream applications.

\subsection{Controllable Generation of Diffusion Models}
Controllable generation for diffusion models has been widely investigated in recent years. One line of work focuses on parameter-efficient specialization: LoRA~\cite{hu2022lora} enables low-rank adaptation with minimal trainable parameters and has become a standard mechanism for style, subject, and domain adaptation in practice. Closely related personalization methods include Textual Inversion~\cite{gal2022textualinversion} and DreamBooth~\cite{ruiz2023dreambooth}, which bind concepts or identities into text-conditioned diffusion pipelines. Another major line introduces explicit conditional pathways. ControlNet~\cite{zhang2023controlnet} attaches trainable control branches to preserve pretrained generation priors while injecting structural constraints such as edges, depth maps, human pose, segmentation, and outline. T2I-Adapter~\cite{mou2023t2iadapter} similarly provides lightweight adapters for condition injection with strong compatibility across downstream tasks. For image-conditioned control, IP-Adapter~\cite{ye2023ipadapter} decouples image and text conditioning to improve identity consistency while retaining textual editability. Our prior work further improves fine-grained and compositional control: AttriCtrl~\cite{chen2025attrictrl} enables continuous intensity control over aesthetic attributes, while EliGen~\cite{zhang2025eligen} introduces entity-level regional attention for precise multi-entity layout and manipulation.

However, most techniques are released as isolated modules with distinct training scripts, parameter formats, and runtime hooks. As a result, combining multiple controls often requires extensive handcrafted engineering, conflict arbitration, and repeated fine-tuning. Our Diffusion Templates framework addresses this gap by treating each control method as a pluggable capability under a unified interface, reducing integration and maintenance cost while preserving composability.

\subsection{Plugin Frameworks for LLMs}
The LLM community has rapidly matured reusable plugin and tool-use paradigms that decouple model intelligence from external capabilities. Early work such as Toolformer~\cite{schick2023toolformer} showed that language models can learn to invoke APIs as part of token-level generation, while ReAct~\cite{yao2023react} demonstrated an effective interleaving of reasoning traces and tool actions. These ideas evolved into practical agent frameworks where planning, tool execution, and memory are composed as modular subsystems~\cite{qin2023toollearning,xi2024agentsurvey}. At the systems layer, interface standardization became increasingly important. Function-calling and tool-calling interfaces in production LLM platforms~\cite{openai2023functioncalling} provide a normalized contract for invoking external tools, and MCP~\cite{anthropic2024mcp} extends this idea toward interoperable context and capability exchange between models and external providers. In parallel, skills and reusable agent components~\cite{anthropic2025skills} reduce duplicated engineering and accelerate capability iteration.

These developments provide a useful design analogy for controllable capabilities in diffusion models. Instead of tightly coupling each controllable generation method to a specific model implementation, one can define stable plugin contracts and capability interfaces. Our framework follows this direction: the diffusion base model acts as a core runtime, and controls are implemented as independent plugins that can be activated, composed, and scheduled within a common framework.

\subsection{KV-Cache as a Capability Interface}
KV-Cache originated as a systems mechanism for avoiding redundant attention computation, but recent LLM research increasingly treats it as a broader runtime abstraction. On the systems side, KV-Cache is central to efficient serving through model inference frameworks~\cite{kwon2023vllm} and attention kernels~\cite{dao2022flashattention,dao2023flashattention2}. Beyond acceleration, several works view KV-Cache as a reusable asset that can be shared, compressed, and resumed across requests: Preble~\cite{srivatsa2024preble} exploits prompt sharing and transferable cache states for long-context or retrieval-heavy workloads, InferCept~\cite{abhyankar2024infercept} preserves KV states across tool interactions, and some studies~\cite{zhang2023h2o,li2024snapkv,qin2024mooncake} further develop cache management, retention, and disaggregated serving around cached model states.

These works suggest an important shift: KV-cache is no longer only an efficiency optimization, but also a practical interface for carrying intermediate capabilities such as reusable context, memory, and resumable execution state. We adopt the same perspective in Diffusion Templates.

\section{Framework Design}

\subsection{Overview}
The Diffusion Templates framework is a unified plugin framework for controllable diffusion generation. It decouples base-model inference from control-capability injection: the base diffusion pipeline remains responsible for generation quality, while external Template models provide reusable control signals through a standardized intermediate interface. Under this design, multiple control capabilities can be activated, composed, and scheduled without repeatedly modifying the internal implementation of each underlying diffusion pipeline.

\begin{figure}[t]
    \centering
    \resizebox{0.95\linewidth}{!}{%
    \begin{tikzpicture}[node distance=6mm and 6mm, >=Latex, font=\footnotesize]
        \node[draw, rounded corners, align=center] (tm1) {Template Model 1};
        \node[draw, rounded corners, below=5mm of tm1, align=center] (tm2) {Template Model 2};
        \node[draw, rounded corners, below=5mm of tm2, align=center] (tm3) {Template Model 3};

        \node[align=center, left=12mm of tm2] (ti) {Template Input};

        \node[right=8mm of tm1, align=center] (tc1) {Template Cache 1};
        \node[right=8mm of tm2, align=center] (tc2) {Template Cache 2};
        \node[right=8mm of tm3, align=center] (tc3) {Template Cache 3};

        \node[right=8mm of tc2, align=center] (tc) {Template Cache};

        \node[draw, dashed, rounded corners, fit=(ti) (tm1) (tm2) (tm3) (tc1) (tc2) (tc3) (tc), inner sep=6mm, label={[align=center]above:Template Pipeline}] (tp) {};

        \node[draw, rounded corners, below=20mm of tc, align=center] (dp) {Diffusion Pipeline};
        \node[left=12mm of dp, align=center] (mi) {Model Input};
        \node[right=12mm of dp, align=center] (mo) {Model Output};

        \draw[->] (ti.east) -- (tm1.west);
        \draw[->] (ti.east) -- (tm2.west);
        \draw[->] (ti.east) -- (tm3.west);

        \draw[->] (tm1.east) -- (tc1.west);
        \draw[->] (tm2.east) -- (tc2.west);
        \draw[->] (tm3.east) -- (tc3.west);

        \draw[->] (tc1.east) -- (tc.west);
        \draw[->] (tc2.east) -- (tc.west);
        \draw[->] (tc3.east) -- (tc.west);

        \draw[->] (mi.east) -- (dp.west);
        \draw[->] (tc.south) -- (dp.north);
        \draw[->] (dp.east) -- (mo.west);
    \end{tikzpicture}
    }
    \caption{Overview of Diffusion Templates framework.}
    \label{fig:template_pipeline_multi}
\end{figure}
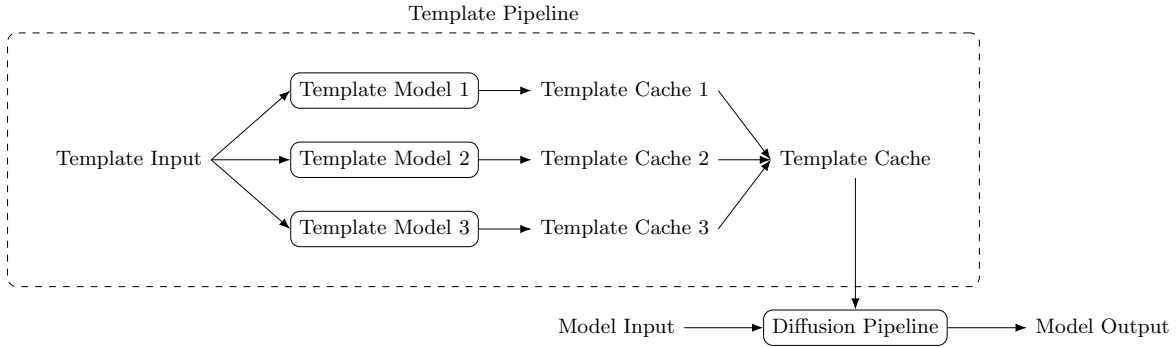

As illustrated in Figure~\ref{fig:template_pipeline_multi}, the framework consists of three core components: \textbf{Template cache}, \textbf{Template model}, and \textbf{Template pipeline}. Template cache serves as the interface for representing model capabilities. A Template model maps arbitrary task-specific inputs to this standardized cache representation, while the input format itself is defined by the corresponding Template model. Template pipeline then orchestrates the loading, execution, and composition of multiple Template models within a unified generation workflow.

\subsection{Template cache}
Unlike LLMs, diffusion models are highly modular and pipeline-centric, usually consisting of components such as text encoders~\cite{radford2021clip,raffel2020t5,yang2025qwen3}, UNet- or DiT-based backbones~\cite{ronneberger2015unet,peebles2023dit}, and a VAE~\cite{kingma2014vae}. Therefore, controllable capabilities cannot be introduced naturally by simply appending textual instructions to the model input. To address this, we define \textbf{Template cache} as a model-capability interface, whose format is constrained to a subset of input arguments accepted by the diffusion pipeline of base models.

This design has two advantages. First, it aligns with existing engineering abstractions of diffusion frameworks, so new capabilities can be integrated by extending pipeline arguments rather than rewriting denoising internals. Second, it provides a stable contract between plugin models and base pipelines, enabling reusable deployment across different downstream tasks.

Among candidate interfaces, KV-Cache is currently the most practical and recommended Template cache type. It offers strong representational capacity, can directly influence generation behavior, and naturally supports sequence-level concatenation, which is particularly important when multiple templates are activated simultaneously. Moreover, exposing KV-Cache through pipeline arguments typically requires only limited framework modification, making it a convenient choice for rapid adaptation to new diffusion backbones. Nevertheless, KV-Cache is only one possible carrier of model capability rather than a restrictive design choice of our framework. Other Template cache formats can also be supported. In particular, lightweight parameterizations such as LoRA can likewise be treated as a form of Template cache for transmitting model capability through the same interface. More broadly, we do not impose a strict restriction on the input format of the diffusion pipeline, thereby preserving the extensibility of the framework as stronger capability interfaces and new architectural abstractions emerge in future work.

\subsection{Template model}
A \textbf{Template model} is any model that maps arbitrary input format to Template cache format. Its architecture is not restricted. A template can be packaged as a local directory or hosted in remote model hubs (e.g., ModelScope\footnote{\url{https://www.modelscope.cn/}} or HuggingFace\footnote{\url{https://huggingface.co/}}). A typical  Template model package contains:
\begin{itemize}
    \item a \texttt{model.py} entry file defining model logic,
    \item an optional \texttt{.safetensors} weight file for parameter storage.
\end{itemize}

To standardize execution and training, each Template model exposes two interfaces:
\begin{itemize}
    \item \texttt{process\_inputs}: no-gradient preprocessing stage for input parsing, feature preparation, and lightweight data transformation;
    \item \texttt{forward}: gradient-related computation stage that produces Template cache outputs for training or inference.
\end{itemize}

This interface split keeps model definition flexible while preserving framework-level compatibility, enabling heterogeneous template architectures to run under one unified runtime.

\subsection{Template pipeline}
In the Template pipeline, given one or multiple enabled Template models, inference proceeds in three steps: (1) run each Template model on its own input to produce Template cache; (2) merge caches according to cache type (e.g., direct concatenation for KV-Cache); (3) pass merged cache into the base diffusion pipeline together with normal generation arguments.

Because KV-Cache natively supports concatenation, multiple Template models can be jointly effective without changing base denoising logic. In practice, template inference can be scheduled in a round-robin manner with lazy loading to reduce peak memory usage when many templates are configured. Importantly, Template models do not enter the base model's denoising loop; they are executed outside the iterative denoising process, so runtime overhead is typically small and inference remains efficient.

\subsection{Template model Training}
The training strategy of a Template model follows the standard paradigm adopted by controllable adaptation methods such as ControlNet and LoRA~\cite{zhang2023controlnet,hu2022lora}. Concretely, we attach trainable side branches to the pretrained base model, keep all base-model parameters frozen, and optimize only the parameters in the newly introduced branches. The optimization objective remains identical to the original pretraining loss of the underlying base model, thereby preserving the learning target while transferring task-specific capability into the Template pathway.

Our training framework is built on DiffSynth-Studio. The standardized \texttt{process\_inputs} and \texttt{forward} interfaces exposed by each Template model enable training to be organized into two stages. In Stage I, input processing is executed in a no-gradient pipeline to produce reusable intermediate features, which can be aggressively cached. In Stage II, optimization is restricted to the gradient-relevant \texttt{forward} path under training objectives defined on Template cache. By decoupling preprocessing from gradient computation, this design reduces redundant computation and improves training efficiency.

\section{Model Zoo}

To evaluate the expressiveness and extensibility of Diffusion Templates, we train a diverse set of Template models on top of FLUX.2-klein-base-4B\footnote{\url{https://huggingface.co/black-forest-labs/FLUX.2-klein-base-4B}}. Unless otherwise specified, all images in this section are generated with a fixed random seed of $0$, a classifier-free guidance scale of $4$~\cite{ho2022cfg}, and $50$ inference steps.

\subsection{Structural Control}
\begin{figure}[p]
    \centering
    \begin{tabular}{>{\centering\arraybackslash}m{0.31\linewidth} >{\centering\arraybackslash}m{0.31\linewidth} >{\centering\arraybackslash}m{0.31\linewidth}}
        Template Input (Depth) & Output 1 & Output 2 \\
        \includegraphics[width=\linewidth]{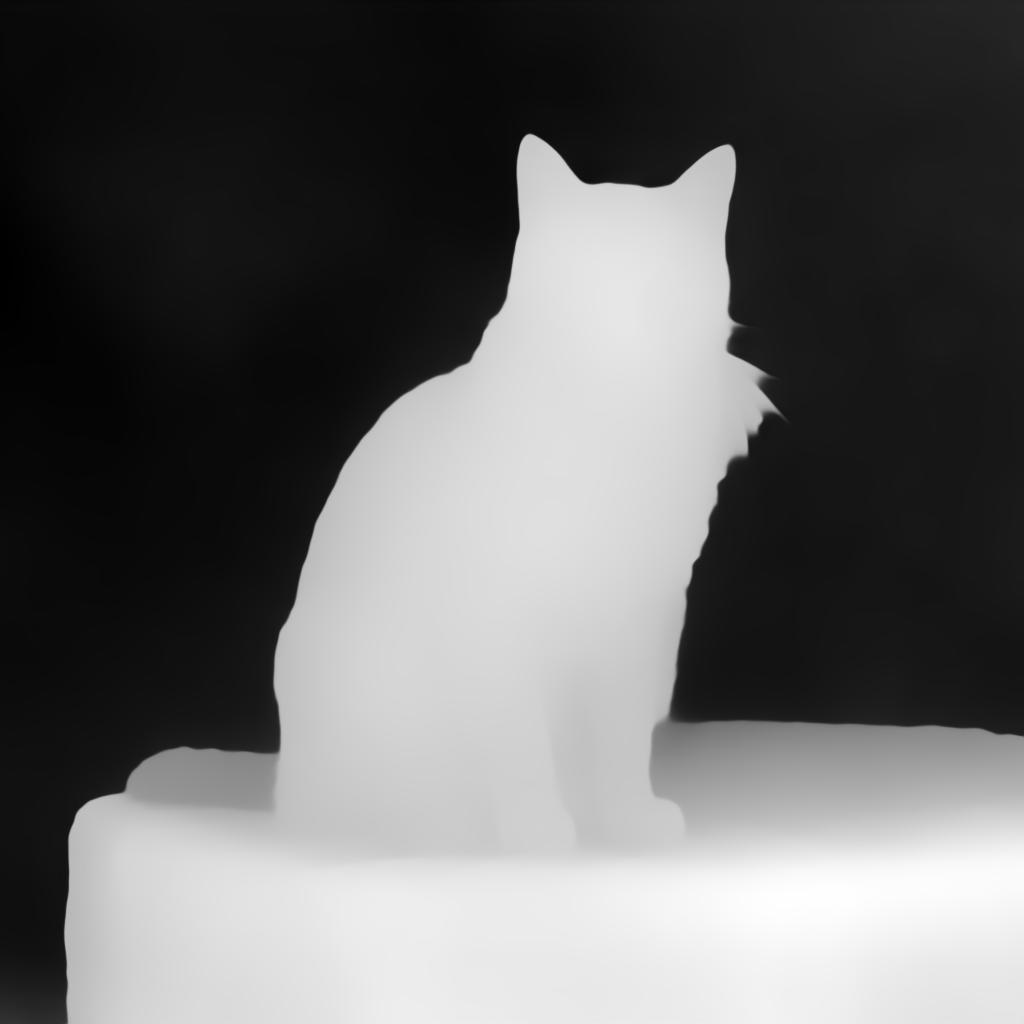} &
        \includegraphics[width=\linewidth]{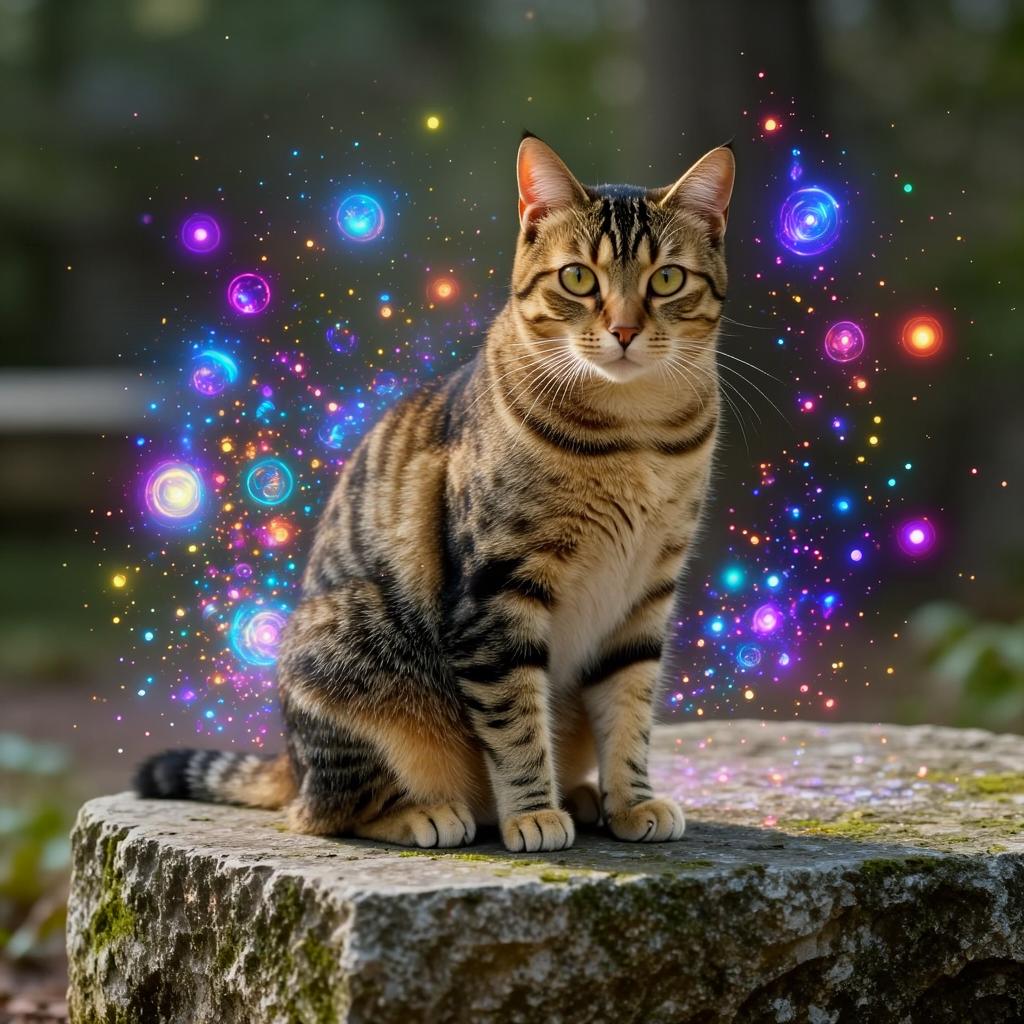} &
        \includegraphics[width=\linewidth]{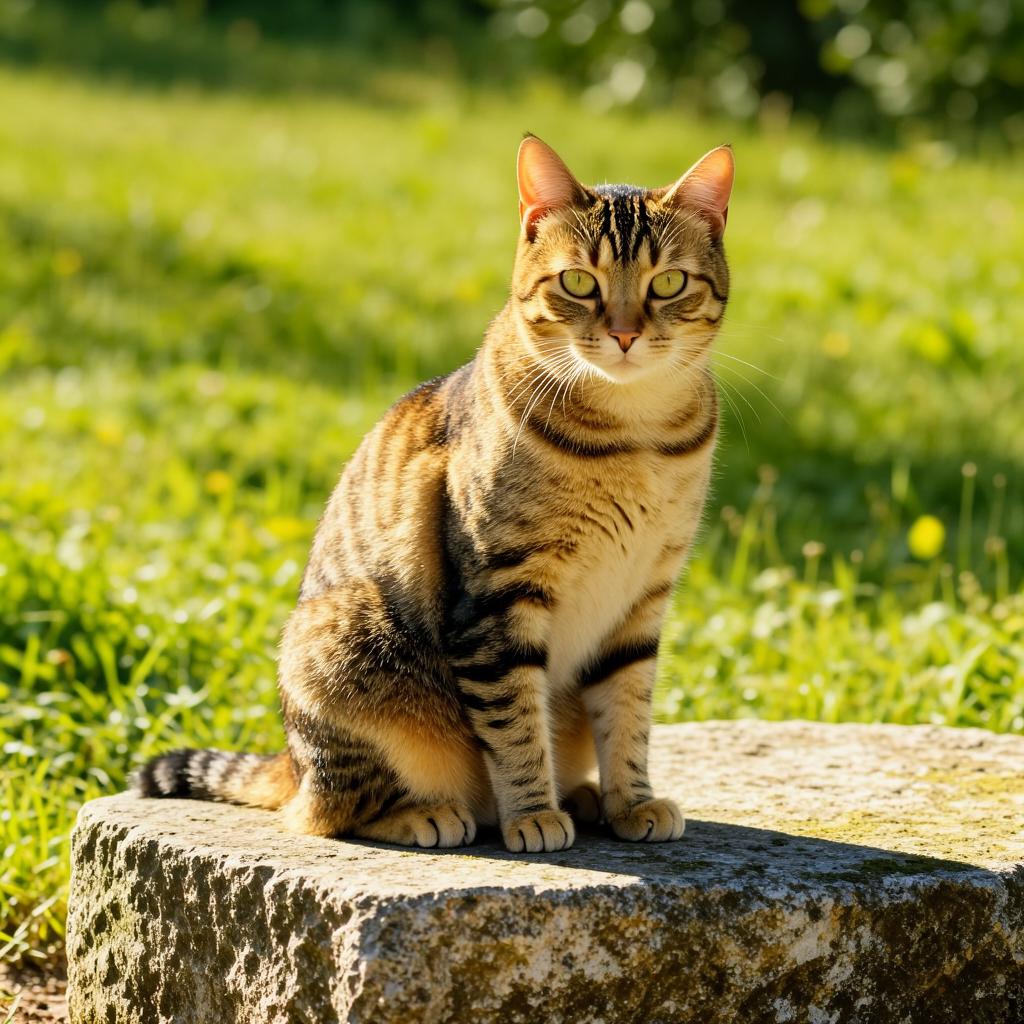} \\
    \end{tabular}
    \caption{Structural control results with a shared depth input. Prompt 1: ``A cat is sitting on a stone, surrounded by colorful magical particles.'' Prompt 2: ``A cat is sitting on a stone, bathed in bright sunshine.''}
    \label{fig:structural_control_depth}
\end{figure}

Structural control was first systematized by ControlNet~\cite{zhang2023controlnet}, which augments a pretrained diffusion model with a trainable branch while preserving the frozen backbone and its generative prior. Following this general idea, we train a structural-control Template model with one key distinction: instead of injecting control signals through residual branches, our method communicates structural information through KV-Cache. The resulting model supports four types of structural conditions, namely depth, outline, human pose, and normal maps. Qualitative depth-conditioned results are shown in Figure~\ref{fig:structural_control_depth}.

\subsection{Brightness Adjustment}
\begin{figure}[p]
    \centering
    \begin{tabular}{>{\centering\arraybackslash}m{0.31\linewidth} >{\centering\arraybackslash}m{0.31\linewidth} >{\centering\arraybackslash}m{0.31\linewidth}}
        Brightness: 0.3 & Brightness: 0.5 & Brightness: 0.7 \\
        \includegraphics[width=\linewidth]{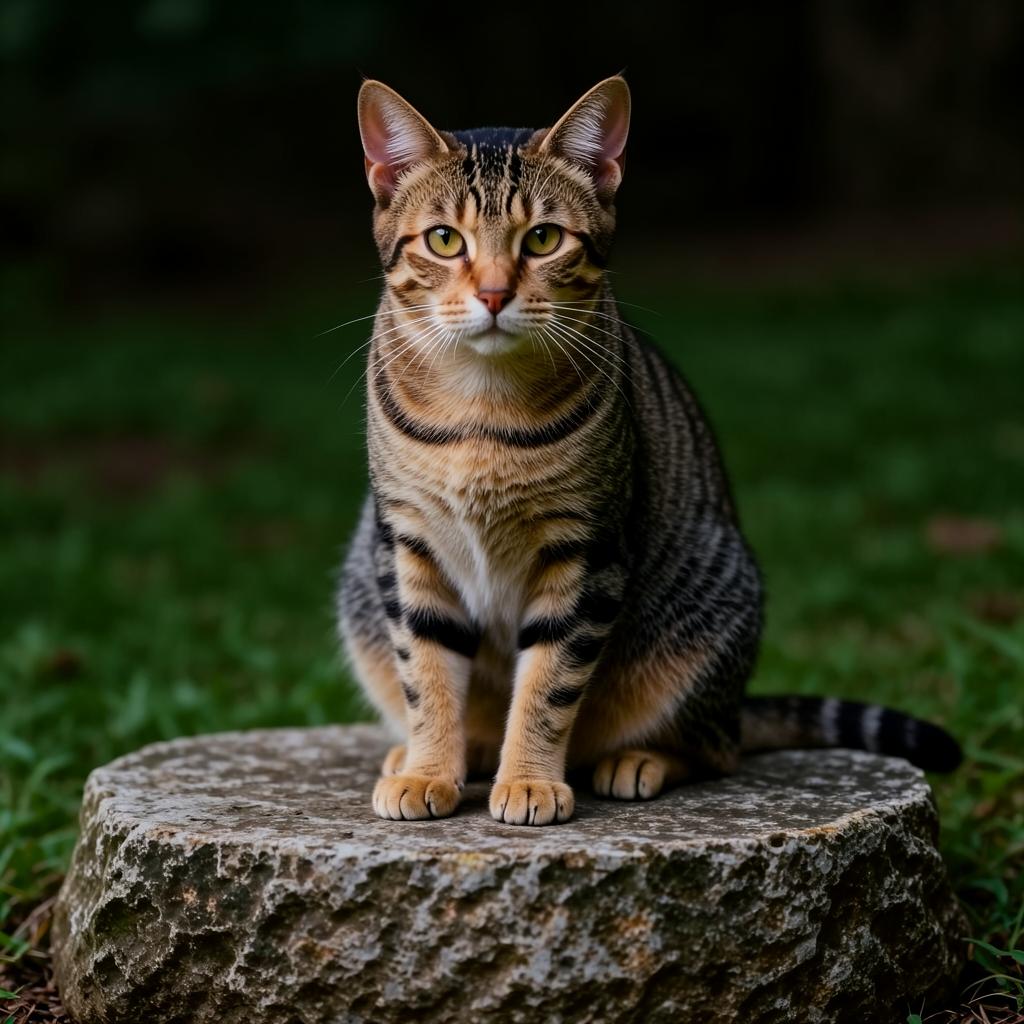} &
        \includegraphics[width=\linewidth]{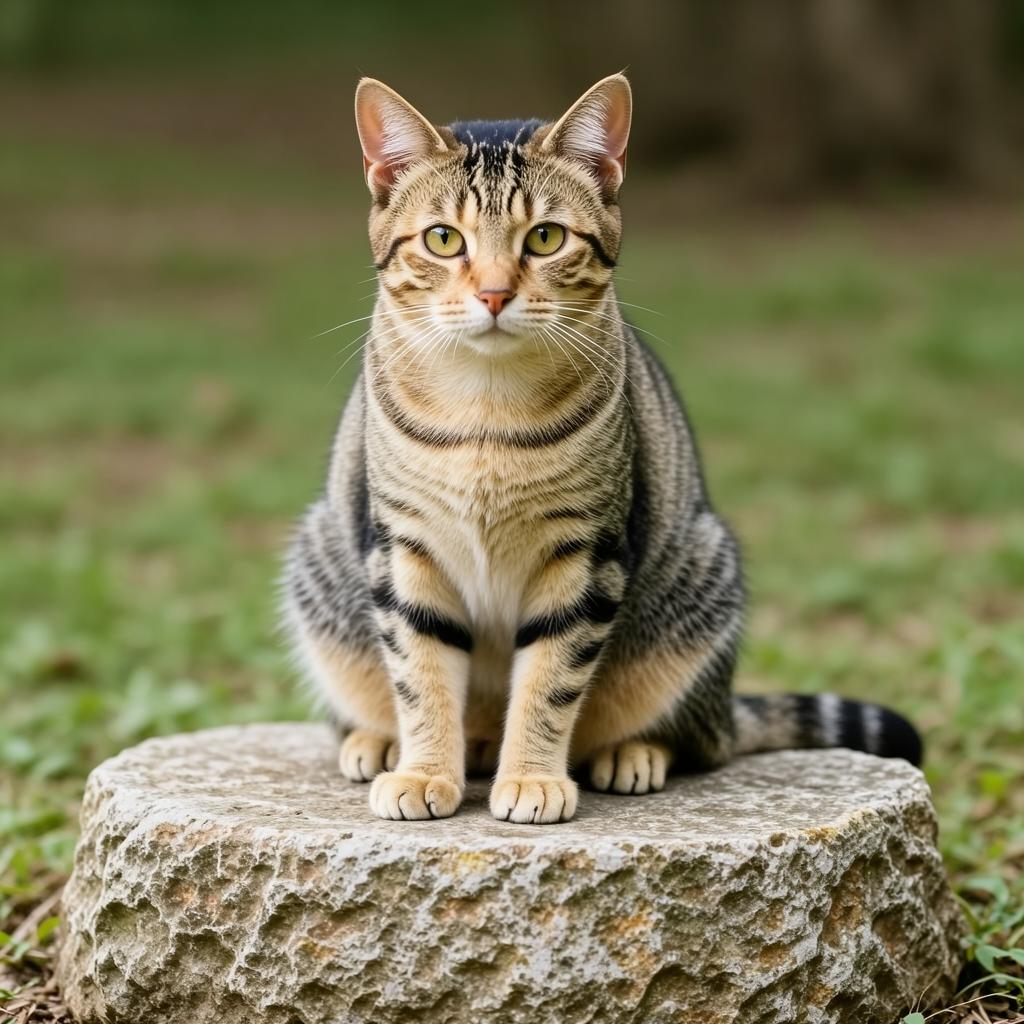} &
        \includegraphics[width=\linewidth]{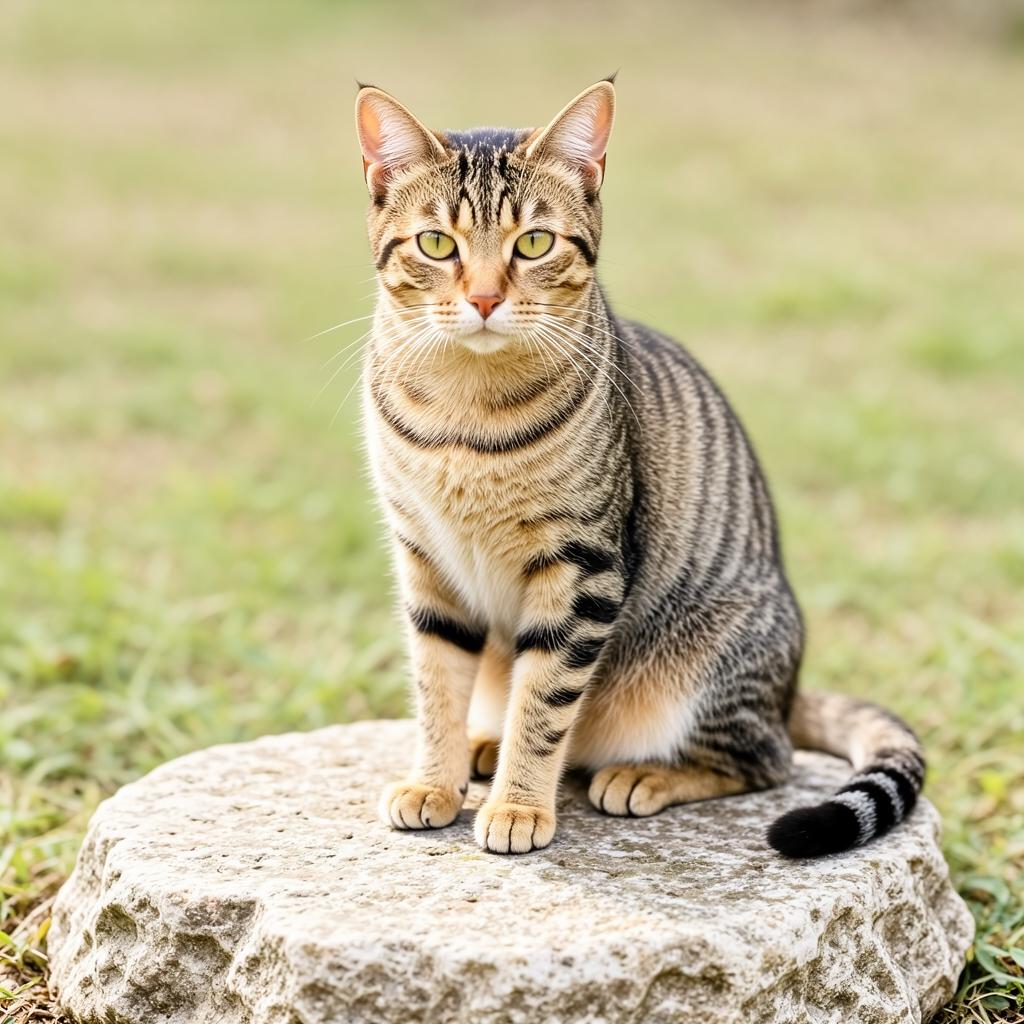} \\
    \end{tabular}
    \caption{Brightness adjustment results with a shared prompt: ``A cat is sitting on a stone.''}
    \label{fig:brightness_adjustment_table}
\end{figure}

A naive approach to brightness control is to directly rescale RGB intensities, but this often leads to visually unnatural results. We therefore train a dedicated brightness-adjustment Template model. Its architecture follows the lightweight design of AttriCtrl~\cite{chen2025attrictrl}, consisting of a positional encoding layer and several fully connected layers. During training, the control input is a scalar brightness value defined as the mean RGB intensity normalized to $[0,1]$. As shown in Figure~\ref{fig:brightness_adjustment_table}, the model adjusts global illumination and scene composition while preserving consistency with the text prompt.

\subsection{Color Adjustment}
\begin{figure}[p]
    \centering
    \definecolor{toneWarm}{HTML}{D0B98A}
    \definecolor{toneNatural}{HTML}{808080}
    \definecolor{toneCool}{HTML}{5EA3AE}
    \begin{tabular}{>{\centering\arraybackslash}m{0.31\linewidth} >{\centering\arraybackslash}m{0.31\linewidth} >{\centering\arraybackslash}m{0.31\linewidth}}
        Color: \#D0B98A (Warm) & Color: \#808080 (Natural) & Color: \#5EA3AE (Cool) \\
        \includegraphics[width=\linewidth]{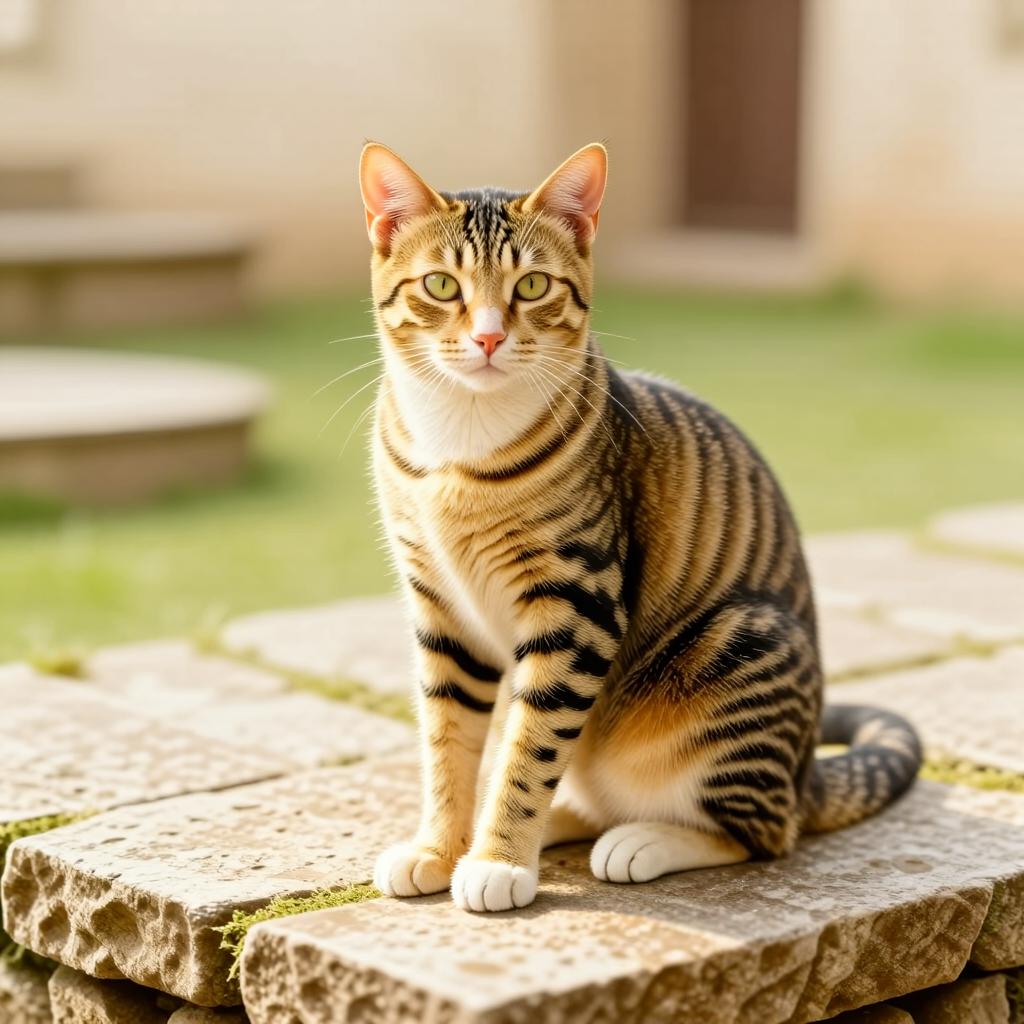} &
        \includegraphics[width=\linewidth]{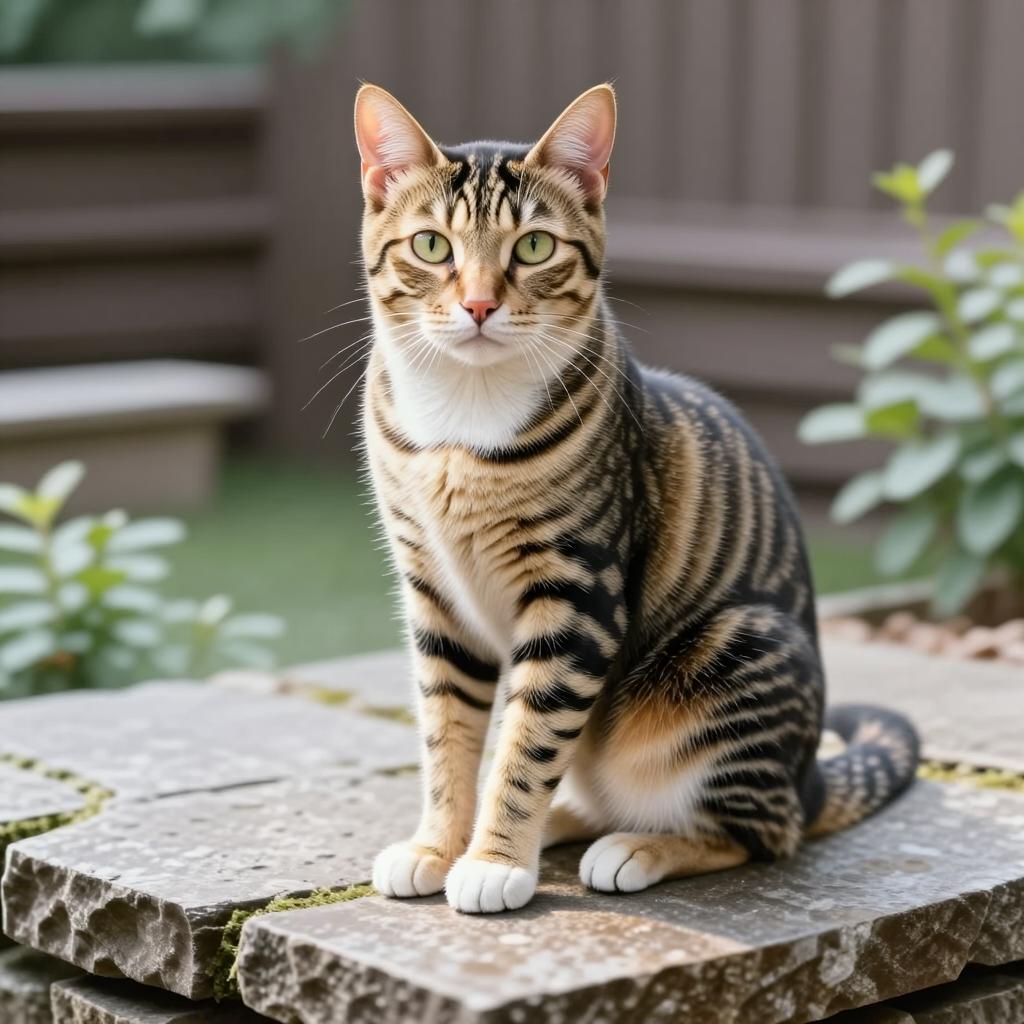} &
        \includegraphics[width=\linewidth]{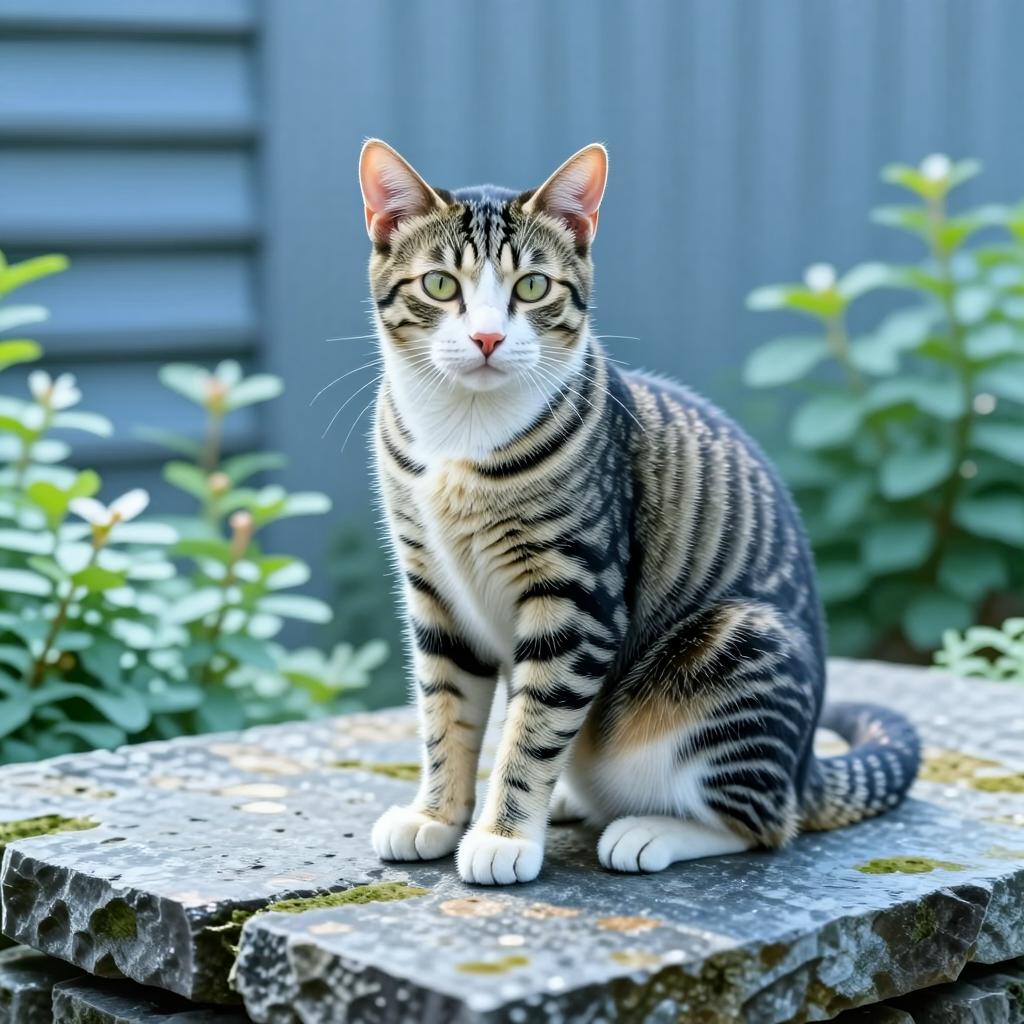} \\
        \fcolorbox{black}{toneWarm}{\rule{0pt}{0.8em}\hspace{2.2cm}} &
        \fcolorbox{black}{toneNatural}{\rule{0pt}{0.8em}\hspace{2.2cm}} &
        \fcolorbox{black}{toneCool}{\rule{0pt}{0.8em}\hspace{2.2cm}} \\
    \end{tabular}
    \caption{Color adjustment results with a shared prompt: ``A cat is sitting on a stone.''}
    \label{fig:tone_adjustment_table}
\end{figure}

Building on the brightness model, we further develop a finer-grained color-adjustment Template model. Instead of a single scalar, this model takes three control inputs corresponding to the mean values of the R, G, and B channels. The training pipeline is otherwise identical to that used for brightness adjustment. Results in Figure~\ref{fig:tone_adjustment_table} show that the control is soft rather than exact: the generated images do not match the target channel values pixel by pixel, but they exhibit a coherent trade-off among color preference, semantic realism, and prompt alignment.

\subsection{Image Editing}
\begin{figure}[p]
    \centering
    \begin{tabular}{>{\centering\arraybackslash}m{0.31\linewidth} >{\centering\arraybackslash}m{0.31\linewidth} >{\centering\arraybackslash}m{0.31\linewidth}}
        Input Image & Output 1 & Output 2 \\
        \includegraphics[width=\linewidth]{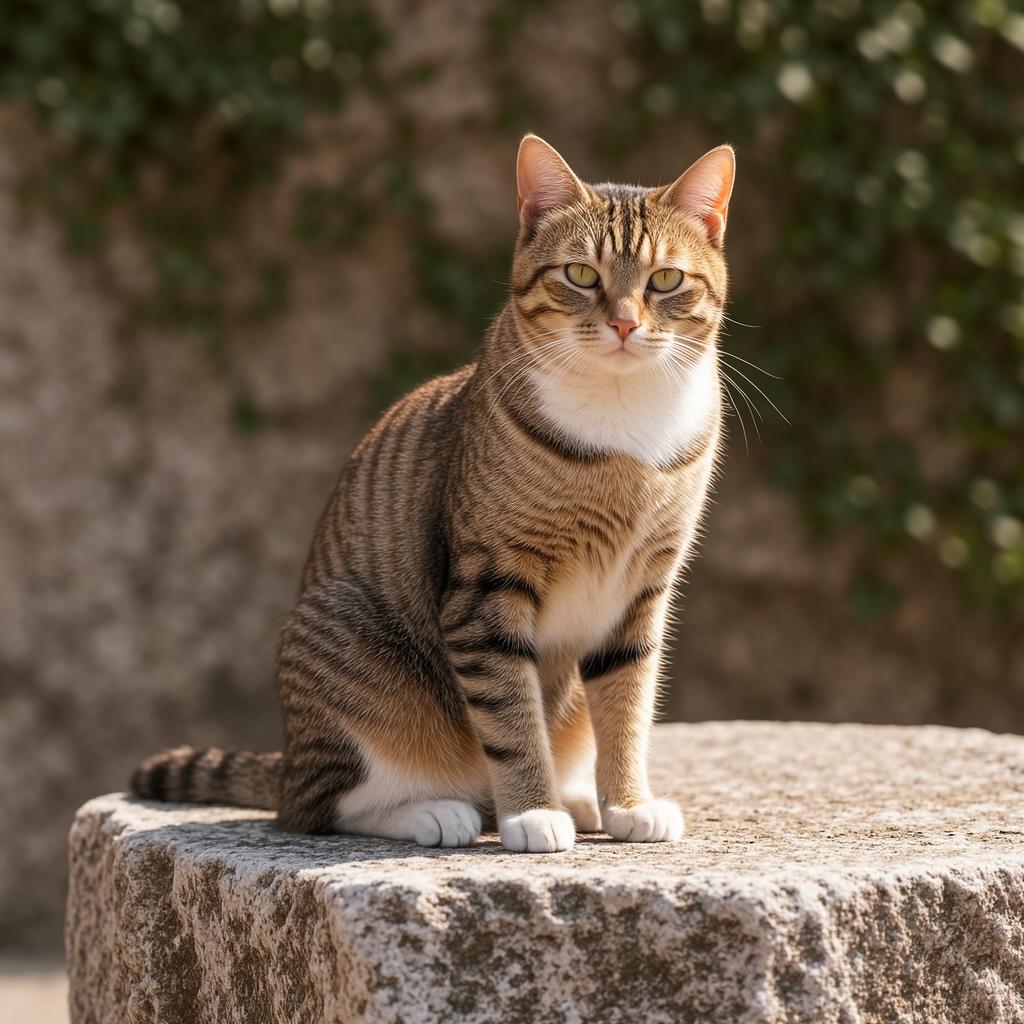} &
        \includegraphics[width=\linewidth]{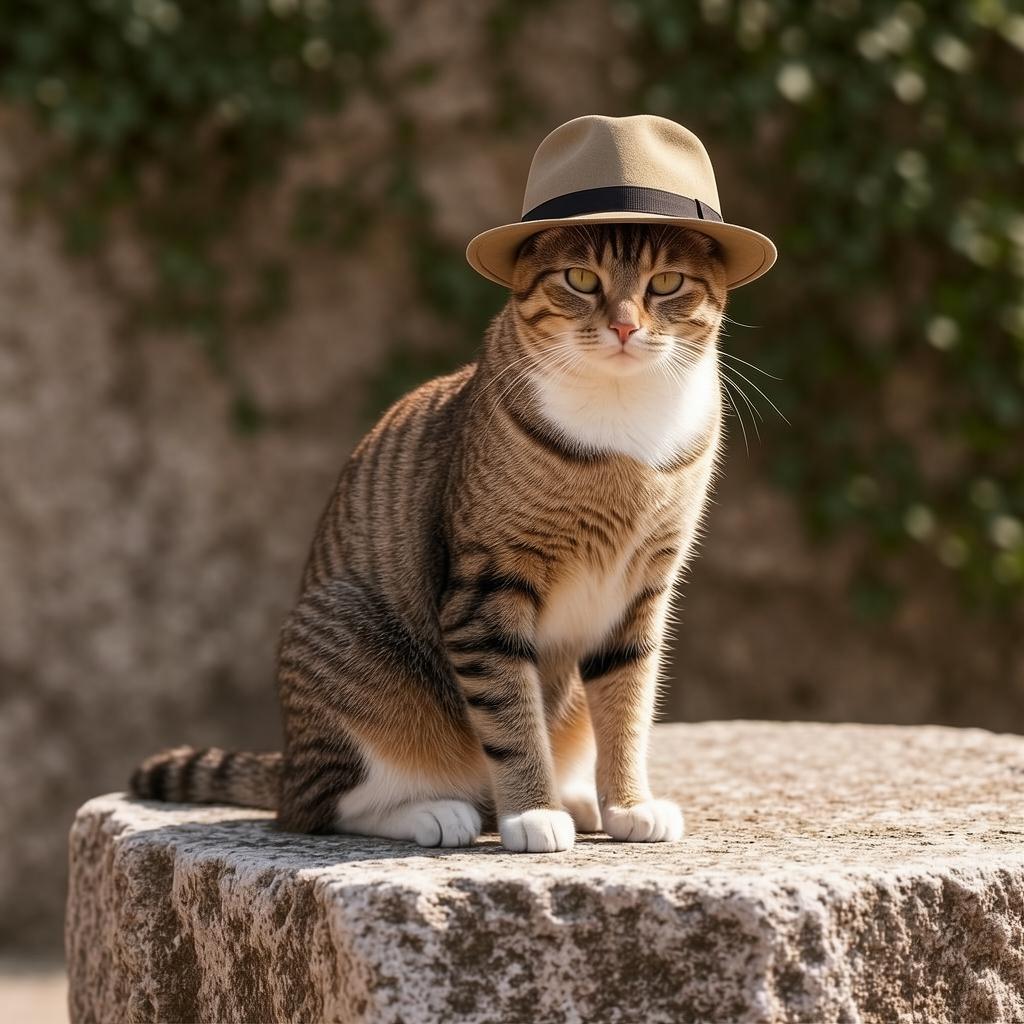} &
        \includegraphics[width=\linewidth]{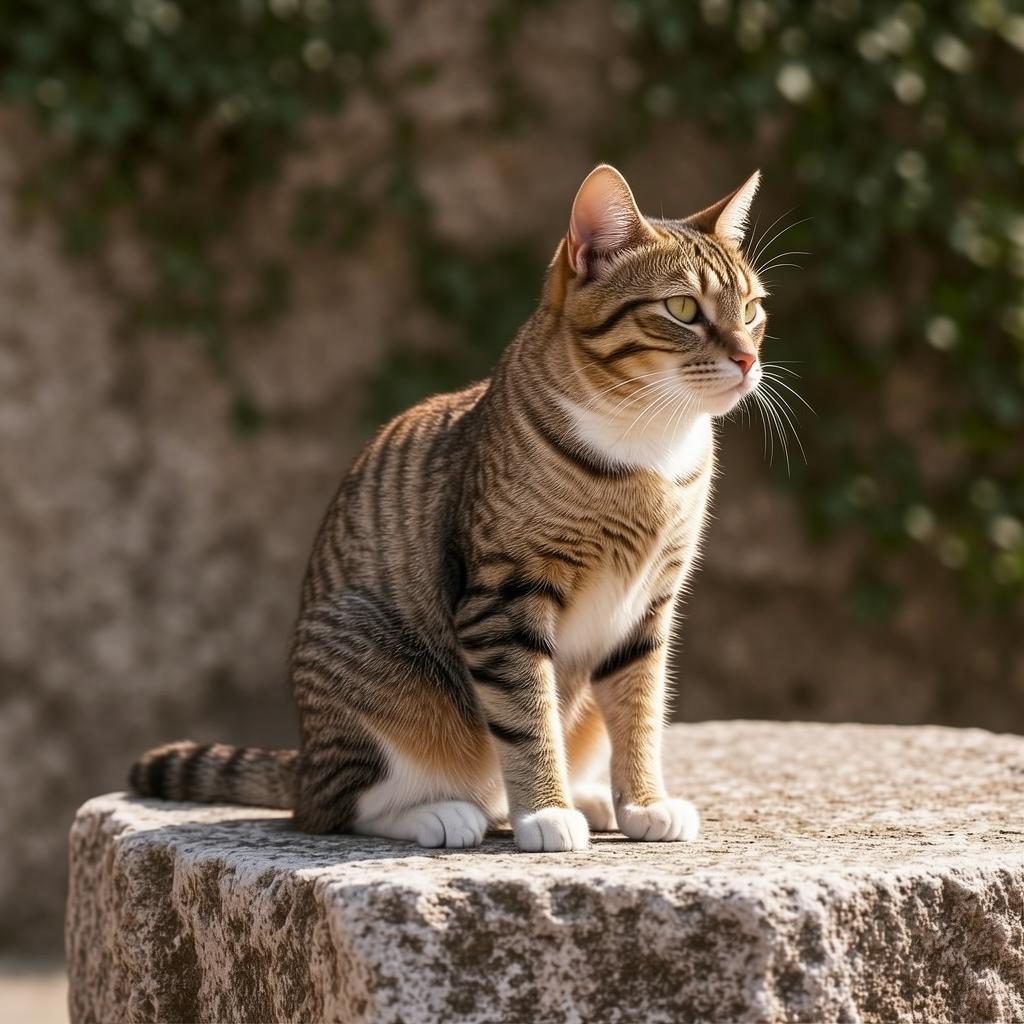} \\
    \end{tabular}
    \caption{Image editing results with a shared reference image. Prompt 1: ``Put a hat on this cat.'' Prompt 2: ``Make the cat turn its head to look to the right.''}
    \label{fig:image_editing_results}
\end{figure}

Although the base model natively supports image editing, editing is substantially more expensive than pure text-to-image generation because of the increased sequence length. We therefore train an image-editing Template model using the same architecture as the structural-control model and transfer the editing capability of the base model into the Template pathway. As shown in Figure~\ref{fig:image_editing_results}, the resulting model achieves editing quality comparable to that of the base model while delivering an empirical inference speedup of approximately 1.8$\times$.

\subsection{Super-Resolution}
\begin{figure}[p]
    \centering
    \begin{tabular}{>{\centering\arraybackslash}m{0.14\linewidth} >{\centering\arraybackslash}m{0.31\linewidth} >{\centering\arraybackslash}m{0.14\linewidth} >{\centering\arraybackslash}m{0.31\linewidth}}
        Input 1 & Output 1 & Input 2 & Output 2 \\
        \includegraphics[width=\linewidth]{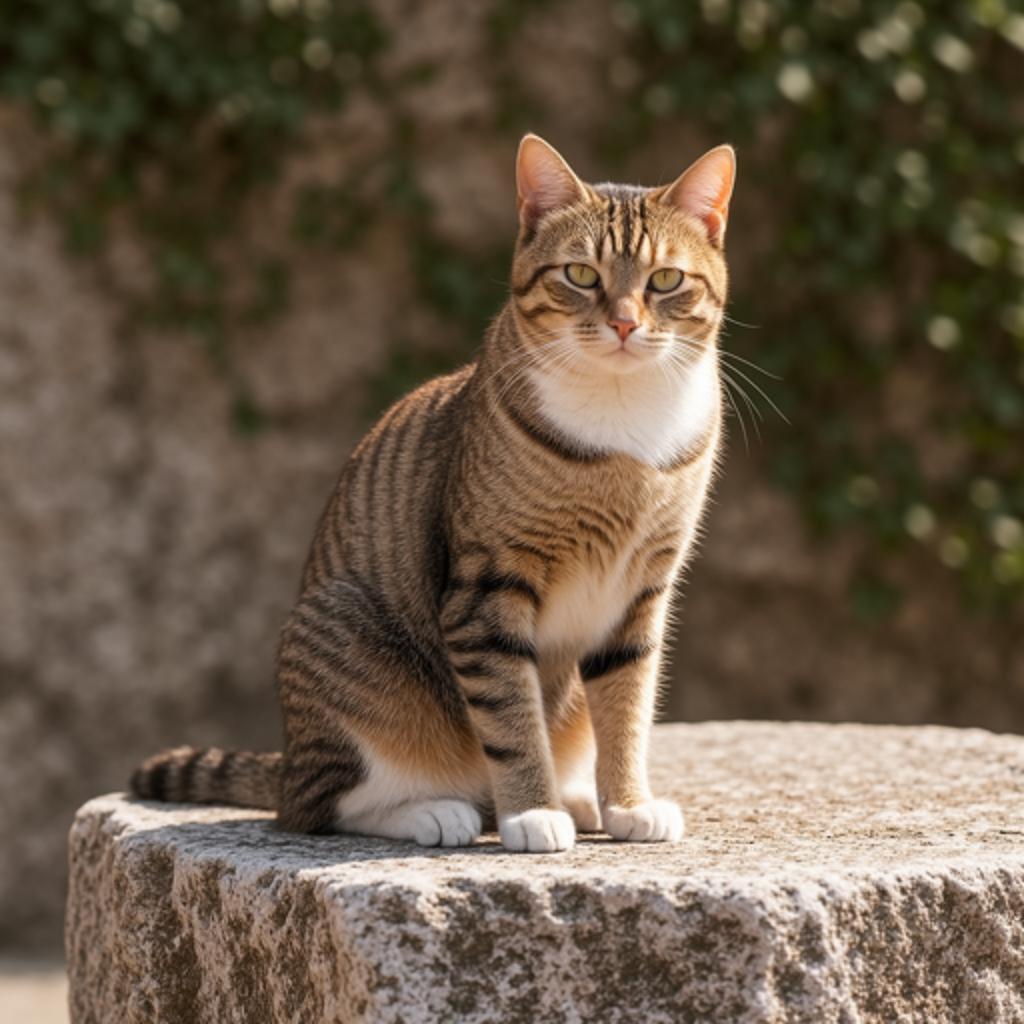} &
        \includegraphics[width=\linewidth]{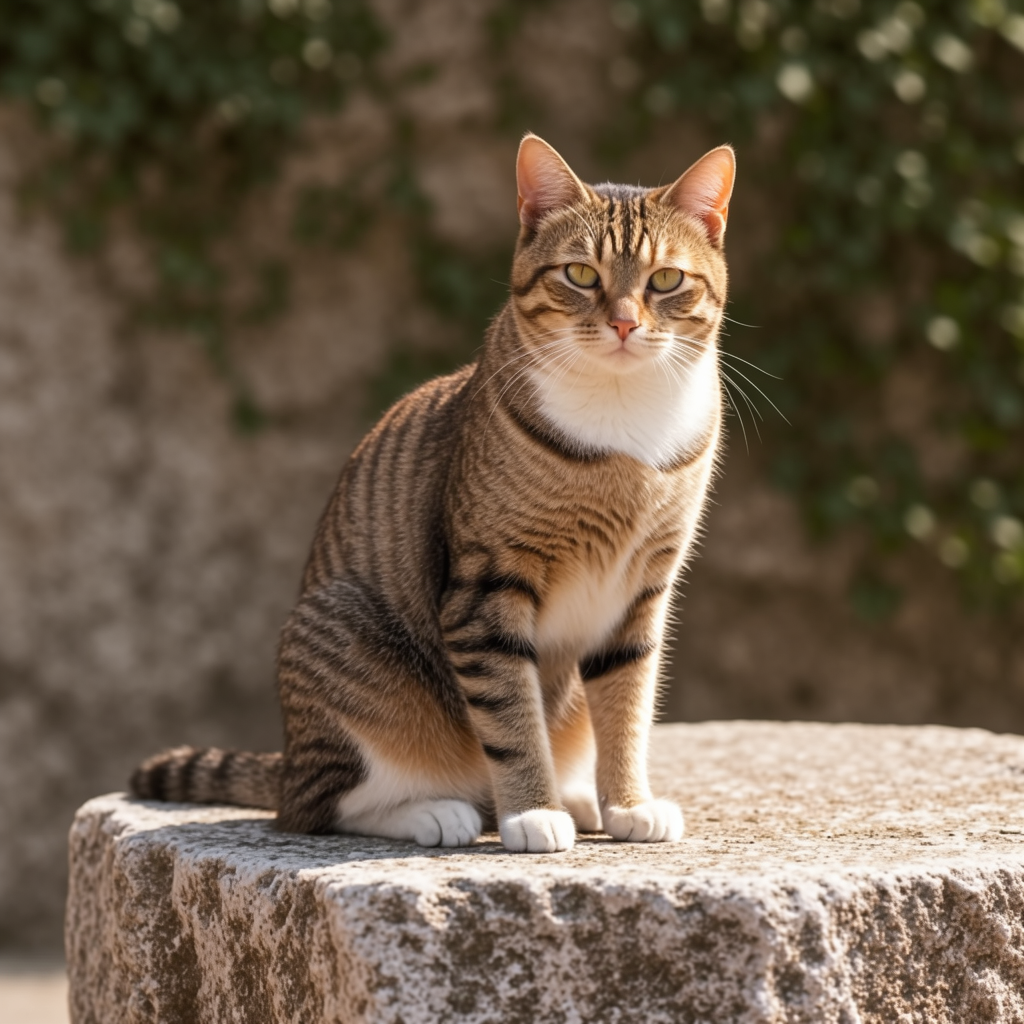} &
        \includegraphics[width=\linewidth]{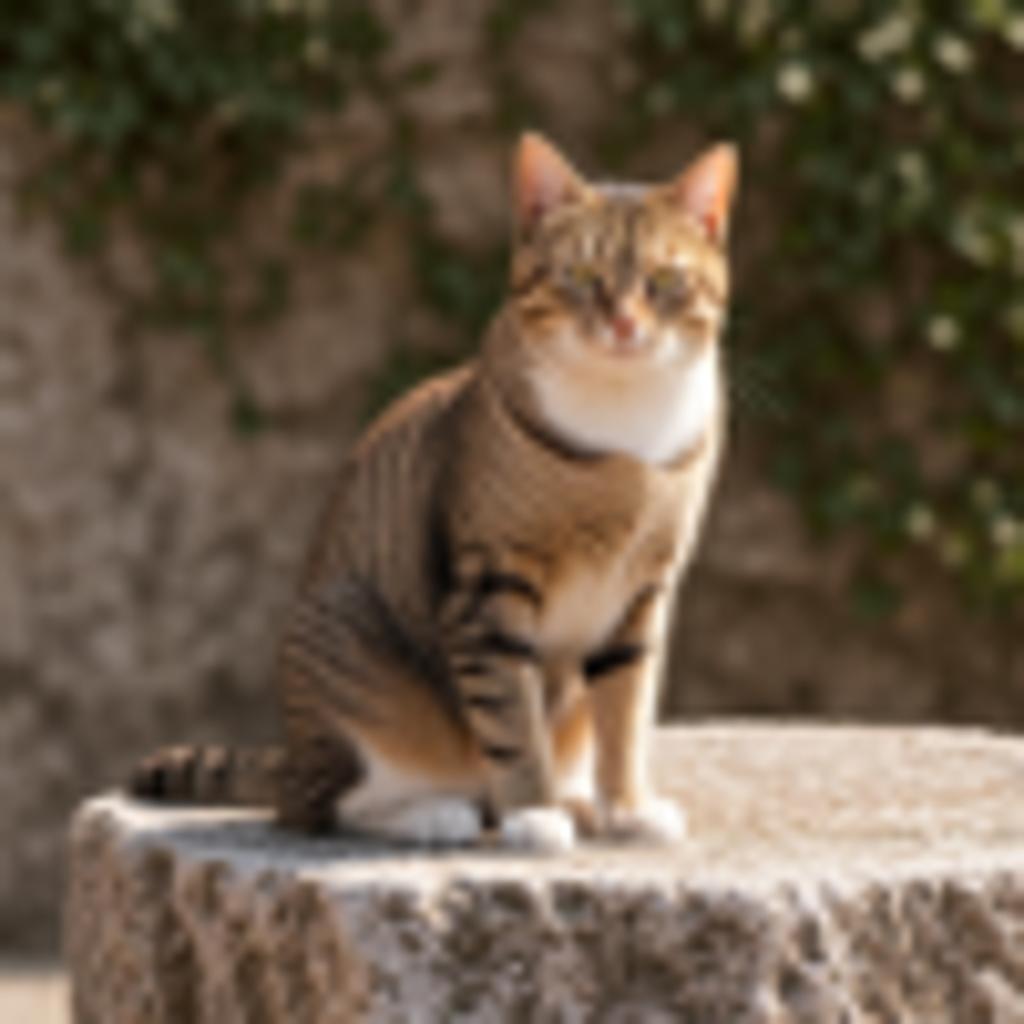} &
        \includegraphics[width=\linewidth]{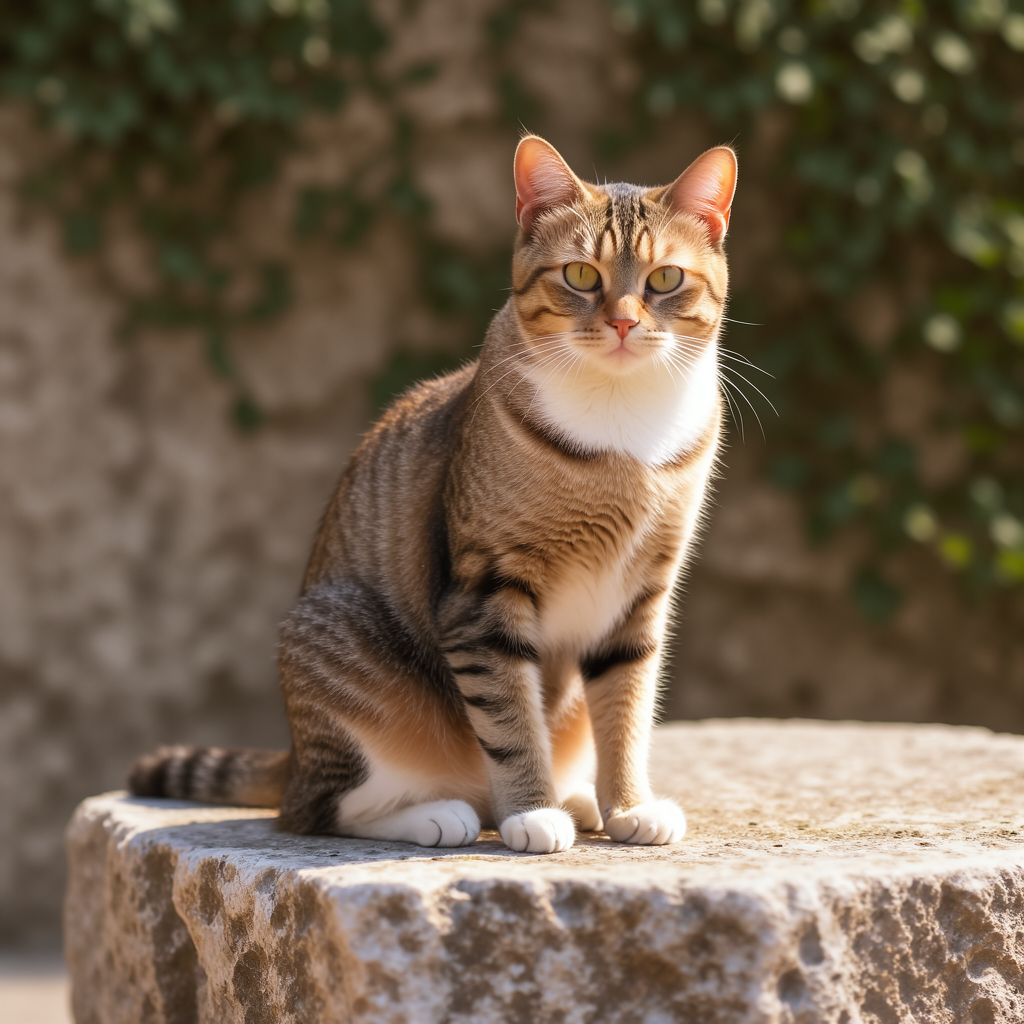} \\
    \end{tabular}
    \caption{Super-resolution results with a shared prompt: ``A cat is sitting on a stone.''}
    \label{fig:super_resolution_outputs}
\end{figure}

Although super-resolution has been extensively studied and specialized models such as Real-ESRGAN are highly effective~\cite{wang2021realesrgan}, we train a super-resolution Template model to evaluate the task coverage of the framework. The architecture is identical to that of the image-editing Template model. Rather than explicitly modeling an upscaling factor, we first bilinearly resize a low-resolution image to the target resolution and then let the Template model recover missing high-frequency details. Figure~\ref{fig:super_resolution_outputs} shows that the model can still produce sharp outputs at large scaling factors, although it remains slower than dedicated super-resolution pipelines.

\subsection{Sharpness Enhancement}
\begin{figure}[p]
    \centering
    \begin{tabular}{>{\centering\arraybackslash}m{0.45\linewidth} >{\centering\arraybackslash}m{0.45\linewidth}}
        Sharpness: 0.1 & Sharpness: 0.8 \\
        \includegraphics[width=\linewidth]{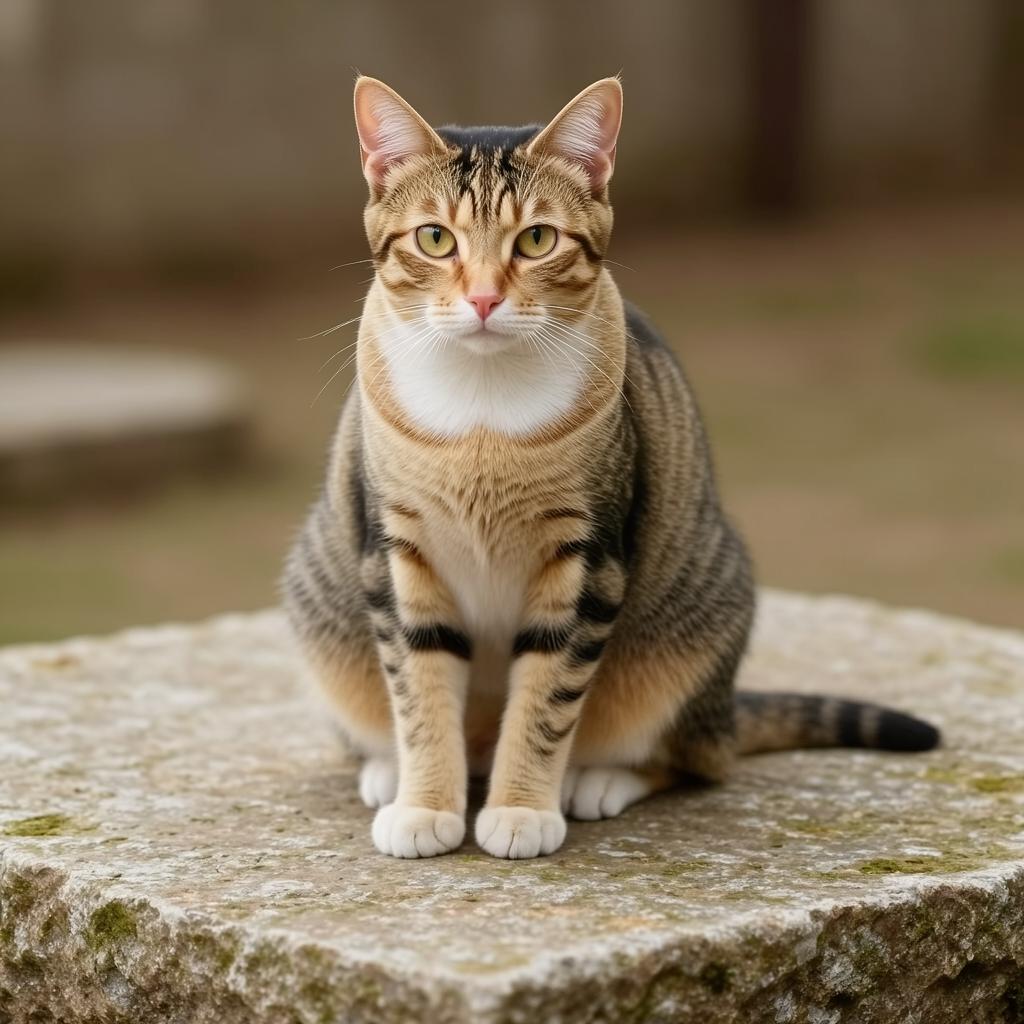} &
        \includegraphics[width=\linewidth]{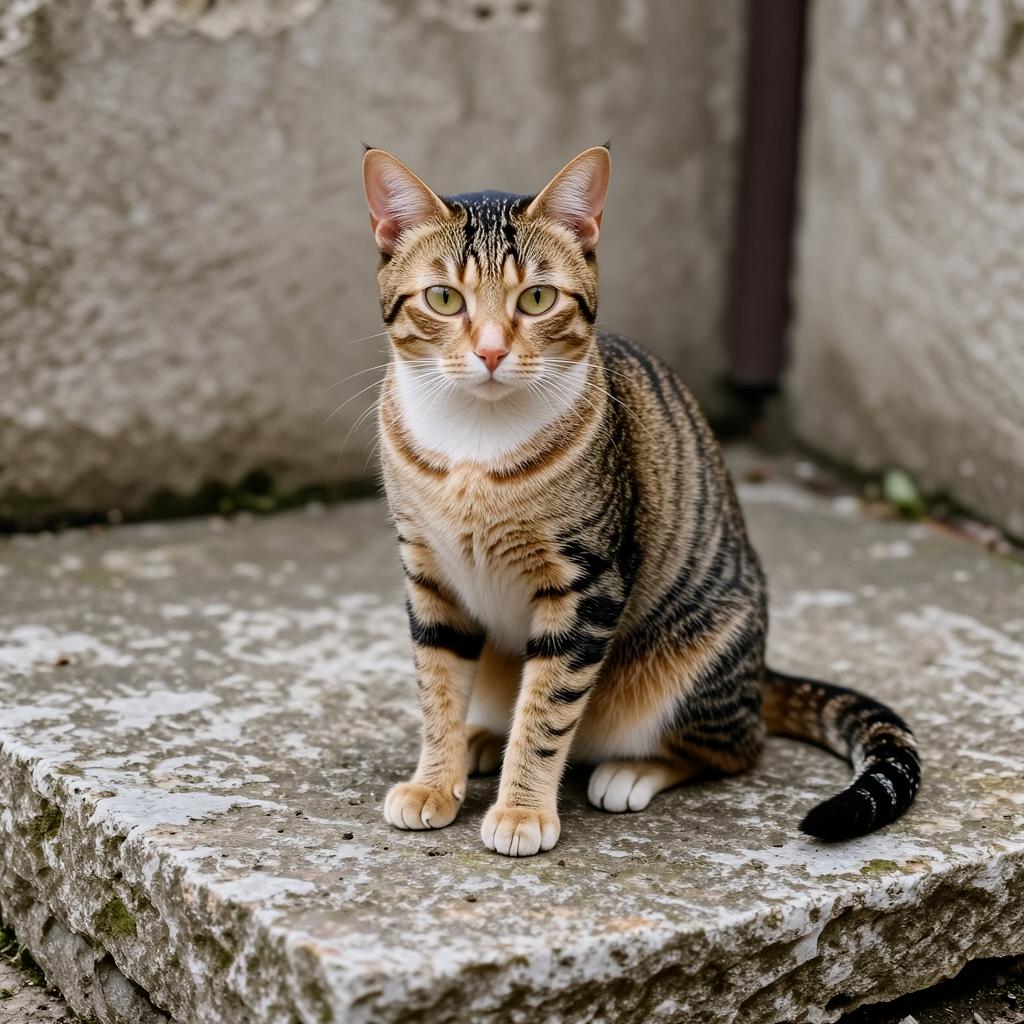} \\
    \end{tabular}
    \caption{Sharpness control results with a shared prompt: ``A cat is sitting on a stone.''}
    \label{fig:sharpness_excitation_table}
\end{figure}

To test whether lightweight Template architectures can control higher-level perceptual attributes, we define a sharpness control signal based on edge density. Specifically, we apply Canny edge detection~\cite{canny1986edge}, compute the fraction of edge pixels, and quantile-normalize this value to $[0,1]$ as the model input. Because sharper images typically contain richer high-frequency boundaries, this statistic serves as a practical proxy for relative sharpness. As shown in Figure~\ref{fig:sharpness_excitation_table}, lower values yield a softer visual appearance, whereas higher values produce clearer structures and stronger local detail.

\subsection{Aesthetic Alignment}
\begin{figure}[p]
    \centering
    \begin{tabular}{>{\centering\arraybackslash}m{0.31\linewidth} >{\centering\arraybackslash}m{0.31\linewidth} >{\centering\arraybackslash}m{0.31\linewidth}}
        Aesthetic scale: 0.0 & Aesthetic scale: 1.0 & Aesthetic scale: 2.5 \\
        \includegraphics[width=\linewidth]{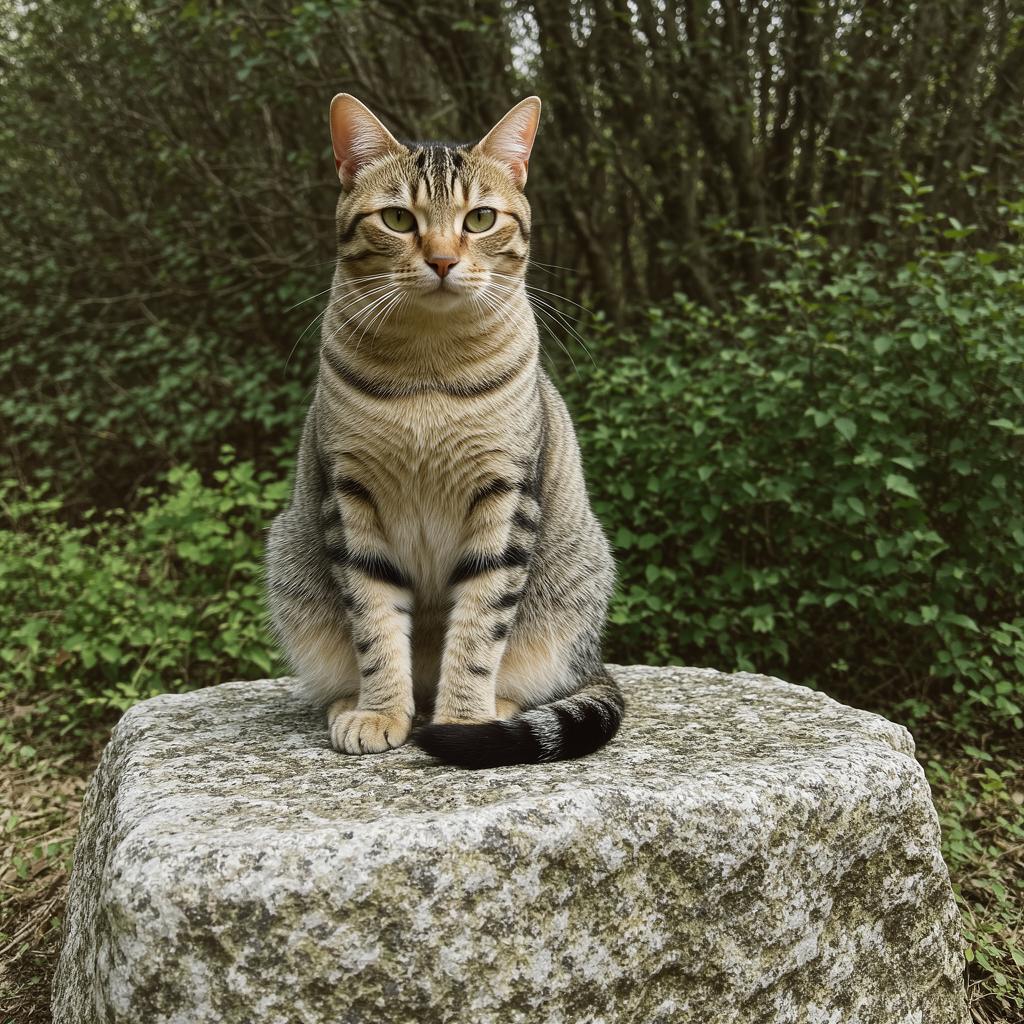} &
        \includegraphics[width=\linewidth]{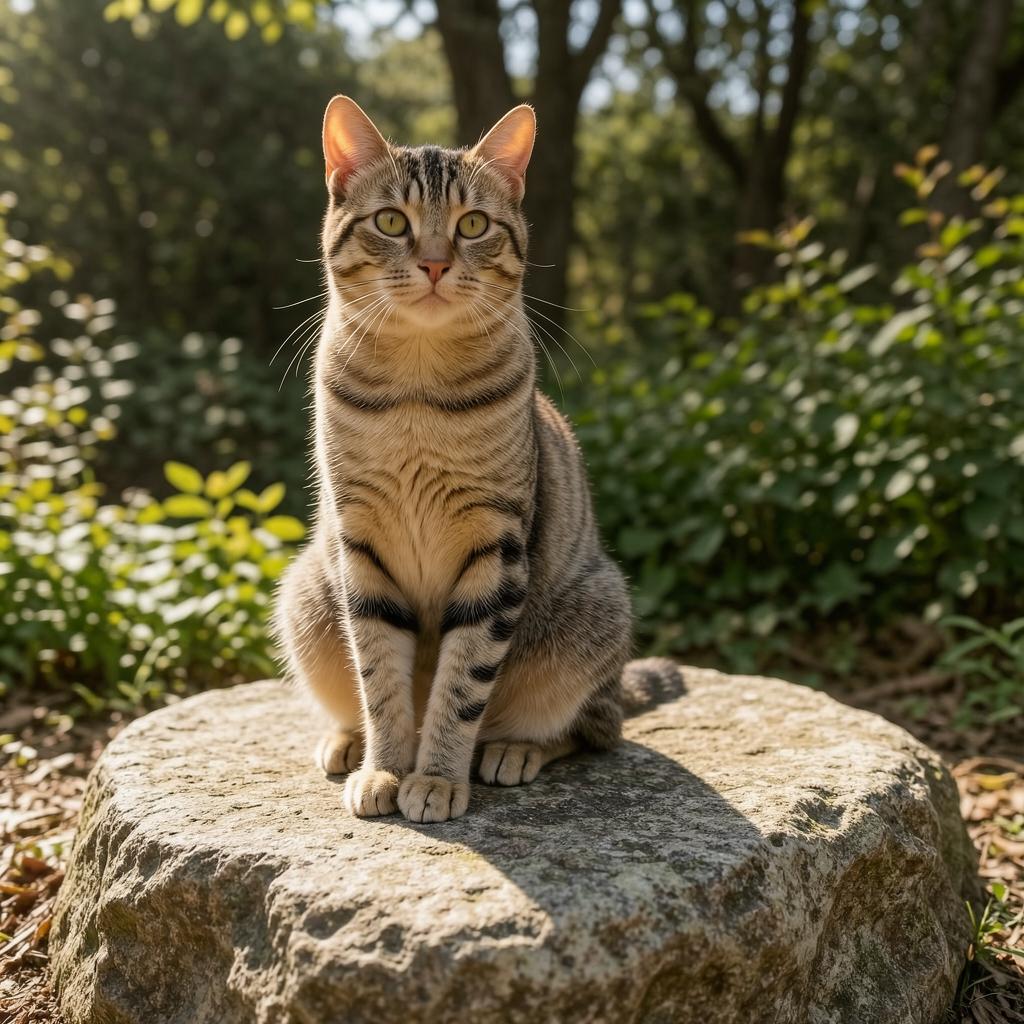} &
        \includegraphics[width=\linewidth]{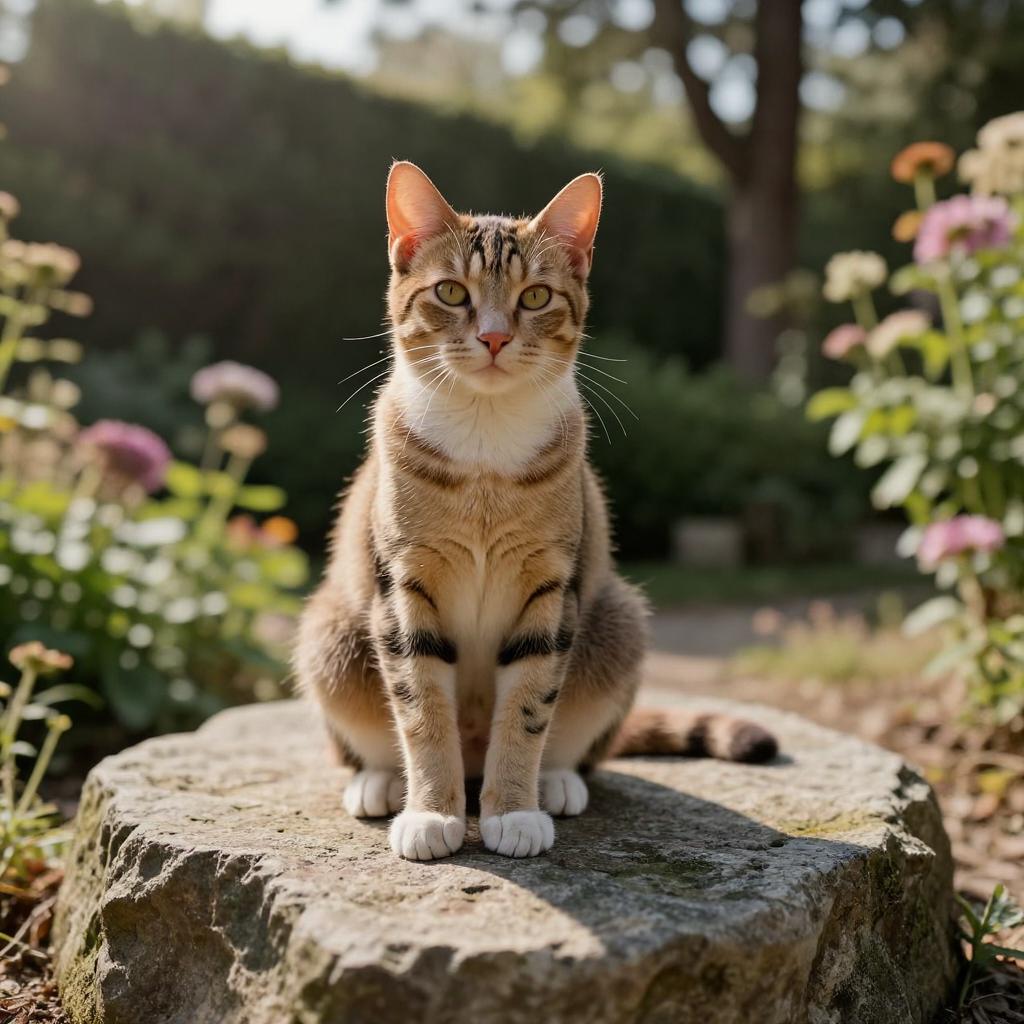} \\
    \end{tabular}
    \caption{Aesthetic alignment results with a shared prompt: ``A cat is sitting on a stone.''}
    \label{fig:aesthetic_alignment_table}
\end{figure}

For scalar control conditions, many image attributes, including brightness, color, and sharpness, can be measured directly from the image itself. Subjective properties such as aesthetics are considerably more challenging, however, because reliable continuous supervision is often unavailable. Studies such as GenAI-Arena~\cite{jiang2024genaiarena} and Pick-a-Pic~\cite{kirstain2023pickapic} instead provide pairwise human preference annotations, in which annotators indicate whether the first image is preferable, the second image is preferable, or the difference is too small to assess confidently. Such supervision is inherently discrete and therefore does not fit naturally into the scalar-control formulation used in the preceding subsections. To address this setting, we adopt LoRA as the capability carrier and interpret it as an input-conditioned parameterization rather than a fixed component of the base model. We construct a small dataset of 90 image pairs generated by the base model, use the preference value to modulate the LoRA strength, and train the corresponding Template model using the differential training strategy of our prior study~\cite{duan2024artaug}. As illustrated in Figure~\ref{fig:aesthetic_alignment_table}, setting the scale to $1.0$ yields softer lighting and a more appealing overall composition. Notably, although the model is trained only on the three values $0$, $0.5$, and $1.0$, it generalizes beyond the training range: increasing the scale to $2.5$ prompts the model to introduce additional decorative elements, such as pink flowers, further enhancing the perceived aesthetic quality. This experiment provides preliminary evidence that Template models can be used for human-preference alignment, and we plan to investigate this direction more systematically in future work.

\subsection{Content Reference}
\begin{figure}[p]
    \centering
    \begin{tabular}{>{\centering\arraybackslash}m{0.14\linewidth} >{\centering\arraybackslash}m{0.31\linewidth} >{\centering\arraybackslash}m{0.14\linewidth} >{\centering\arraybackslash}m{0.31\linewidth}}
        Input 1 & Output 1 & Input 2 & Output 2 \\
        \includegraphics[width=\linewidth]{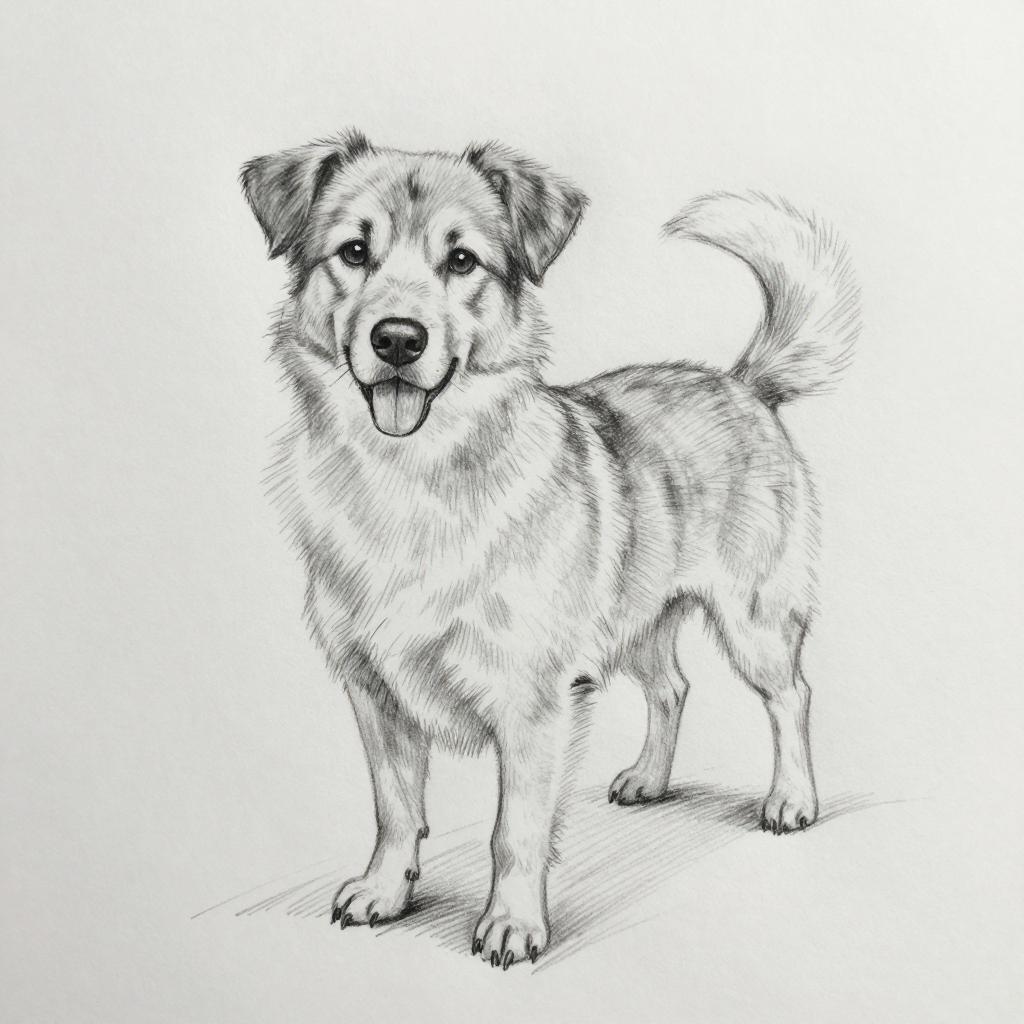} &
        \includegraphics[width=\linewidth]{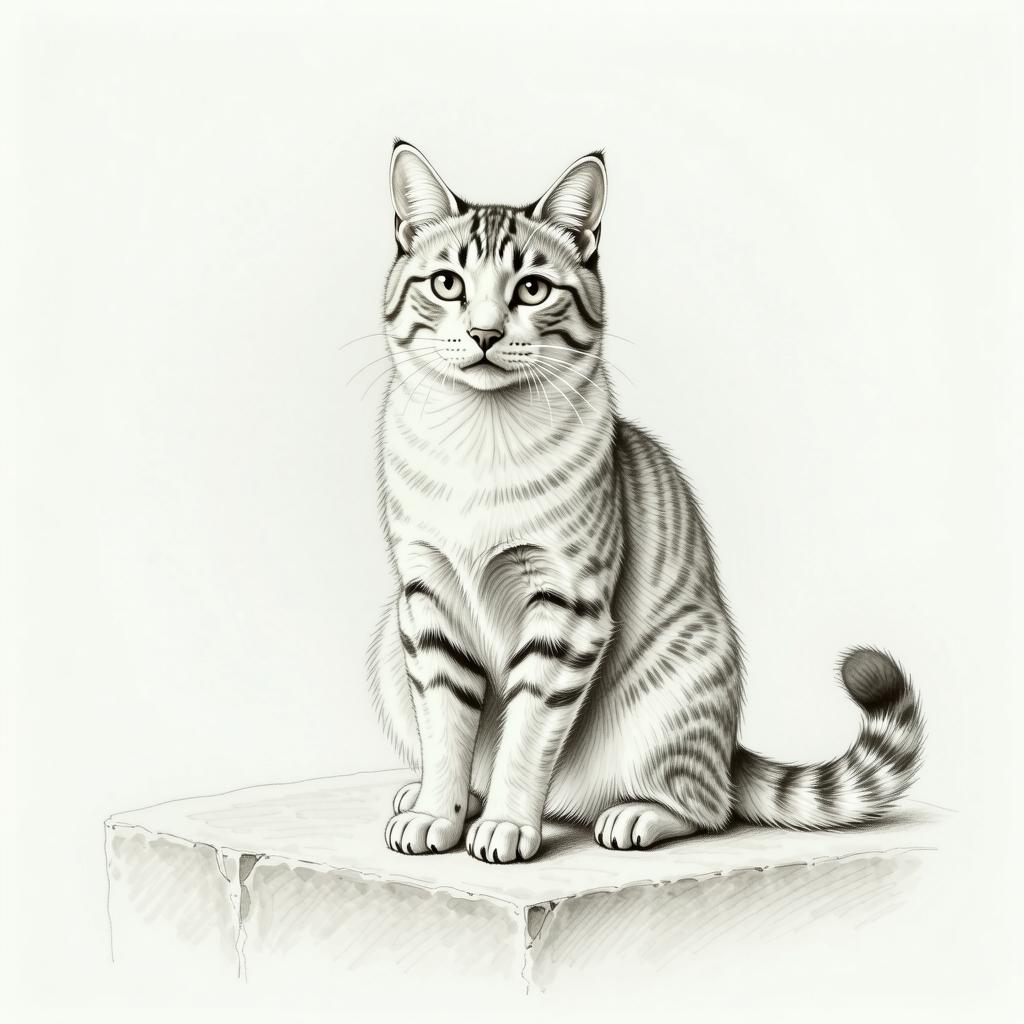} &
        \includegraphics[width=\linewidth]{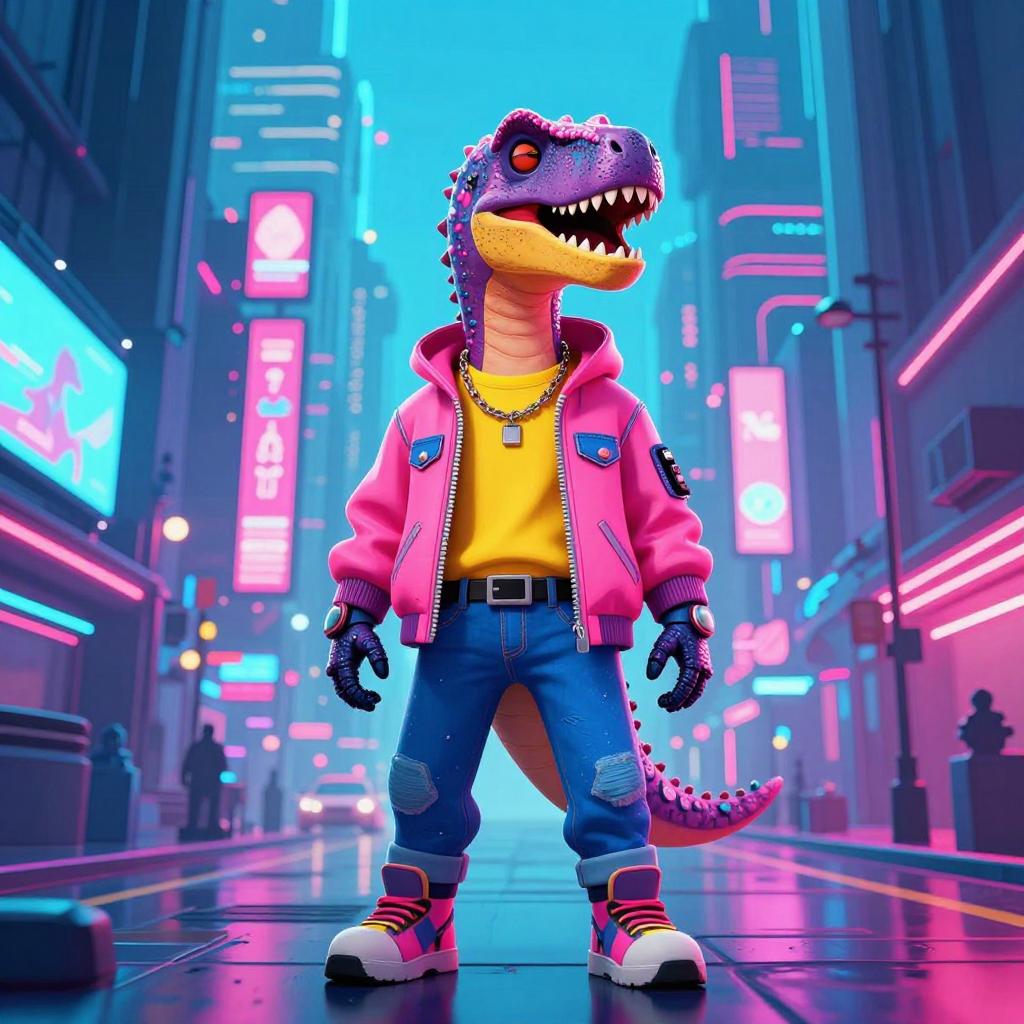} &
        \includegraphics[width=\linewidth]{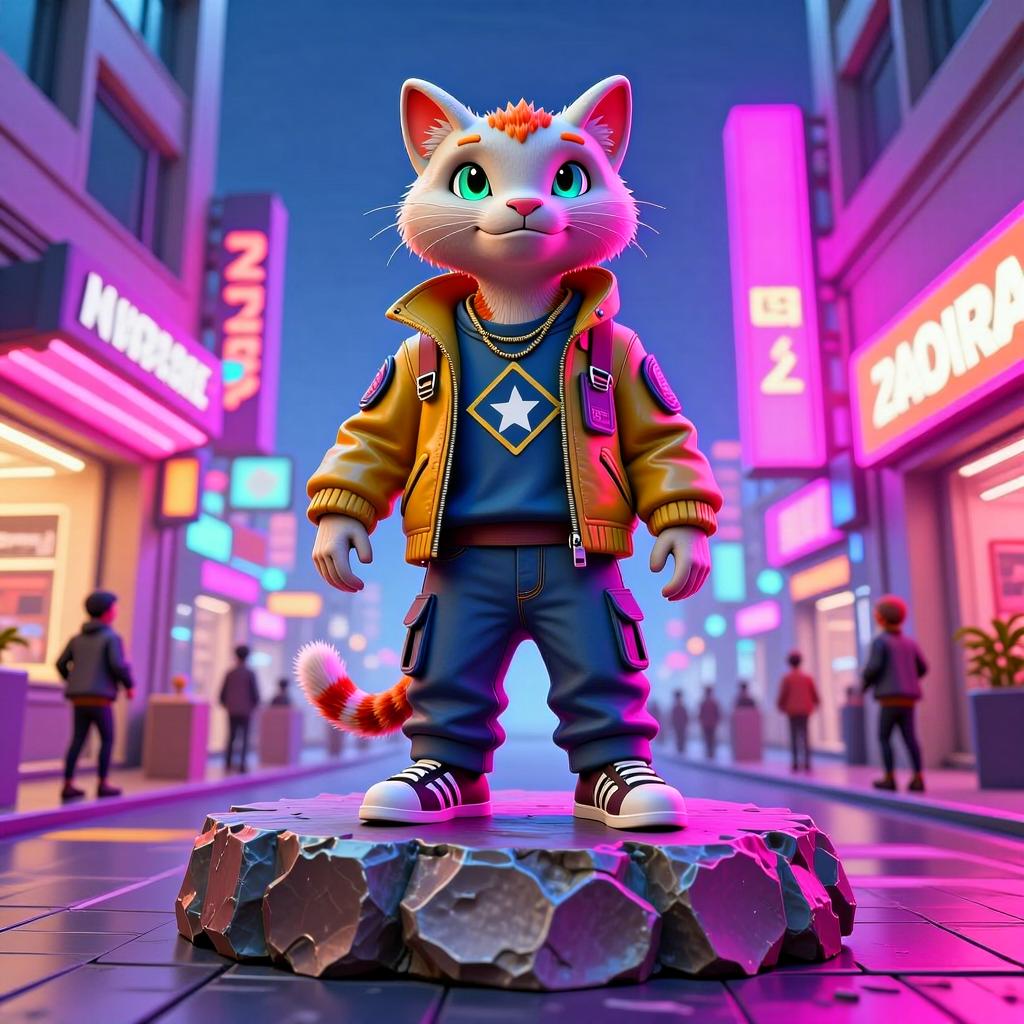} \\
    \end{tabular}
    \caption{Content-reference results with a shared prompt: ``A cat is sitting on a stone.''}
    \label{fig:content_reference_table}
\end{figure}

Building on the aesthetic-alignment experiment, which demonstrates the feasibility of using LoRA as a carrier of model capability, we further develop an Image-to-LoRA Template model for content reference. The model employs SigLIP2~\cite{tschannen2025siglip2} as the image encoder, and maps the resulting visual representation to LoRA weights through several fully connected layers. We train this model on an image-text paired dataset. This formulation is particularly interesting because it enables a reference image to be converted directly into a LoRA representation, which can then be injected into the generation pipeline to produce a new image conditioned on information extracted from the reference. At the same time, the specific content transferred from the reference image is not explicitly controllable. As shown in Figure~\ref{fig:content_reference_table}, the model may in some cases primarily inherit the global visual style of the input image, while in other cases it may instead preserve more concrete attributes, such as character pose and clothing. These observations suggest that this model family exhibits a distinctive and flexible form of reference-based generation, with substantial room for further exploration.

\subsection{Local Inpainting}
\begin{figure}[p]
    \centering
    \begin{tabular}{>{\centering\arraybackslash}m{0.14\linewidth} >{\centering\arraybackslash}m{0.31\linewidth} >{\centering\arraybackslash}m{0.14\linewidth} >{\centering\arraybackslash}m{0.31\linewidth}}
        Input 1 & Output 1 & Input 2 & Output 2 \\
        \parbox[c]{\linewidth}{\centering
            \includegraphics[width=\linewidth]{assets/image_reference.jpg} \\[2.0ex]
            \includegraphics[width=\linewidth]{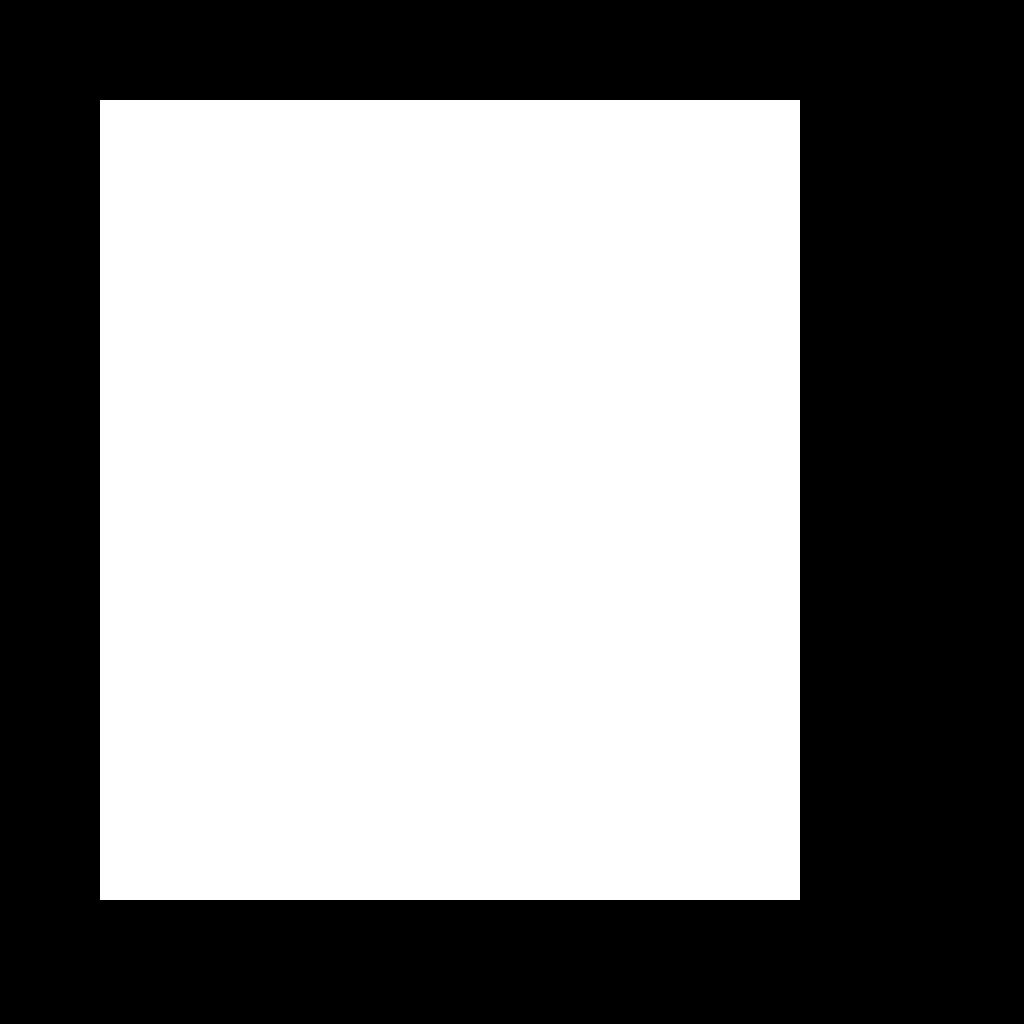}
        } &
        \parbox[c]{\linewidth}{\centering
            \includegraphics[width=\linewidth]{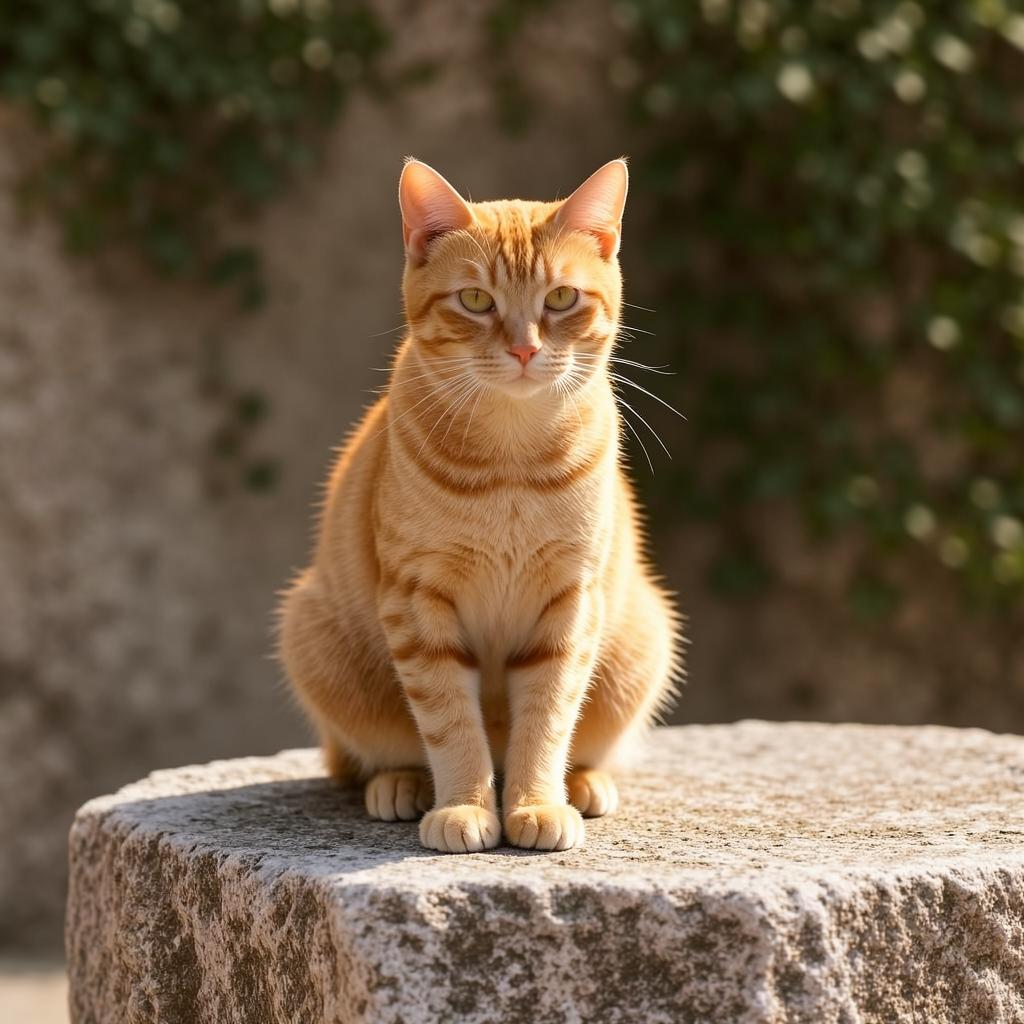}
        } &
        \parbox[c]{\linewidth}{\centering
            \includegraphics[width=\linewidth]{assets/image_reference.jpg} \\[2.0ex]
            \includegraphics[width=\linewidth]{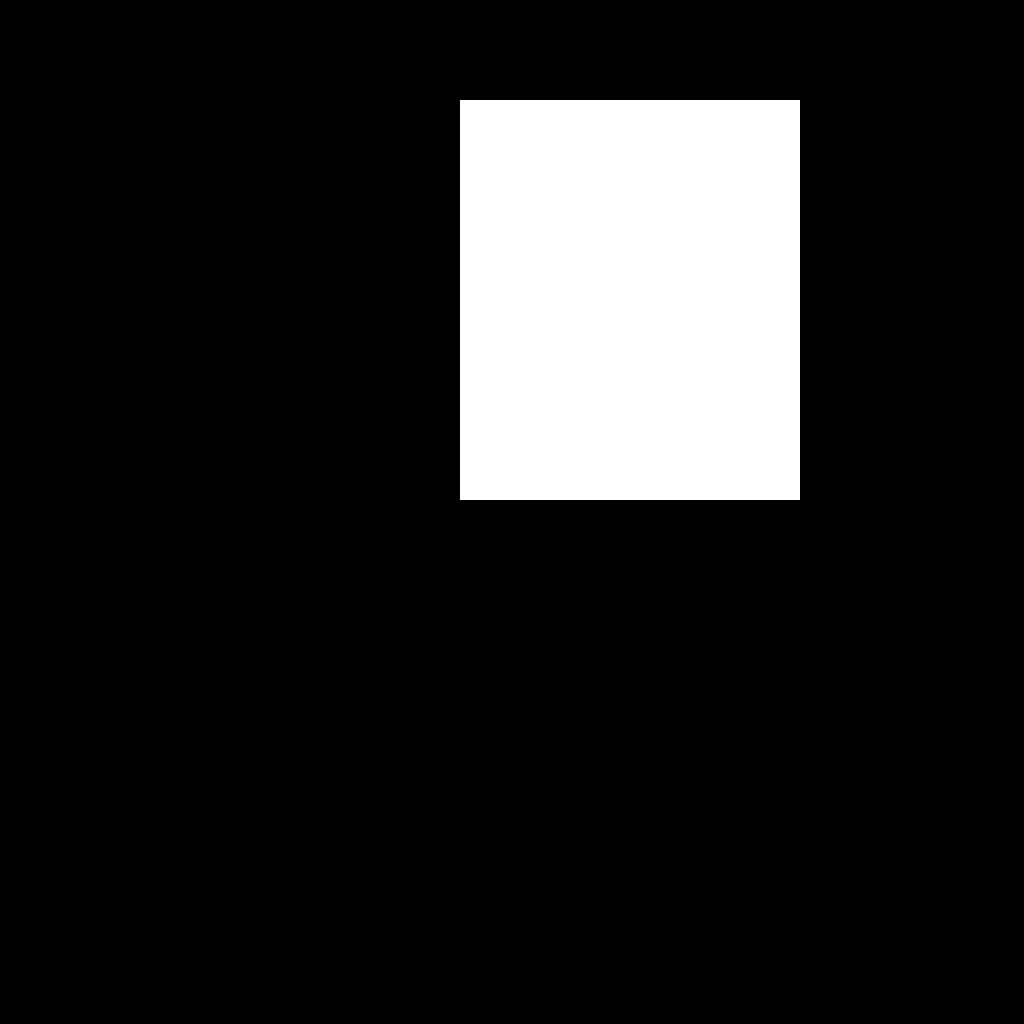}
        } &
        \parbox[c]{\linewidth}{\centering
            \includegraphics[width=\linewidth]{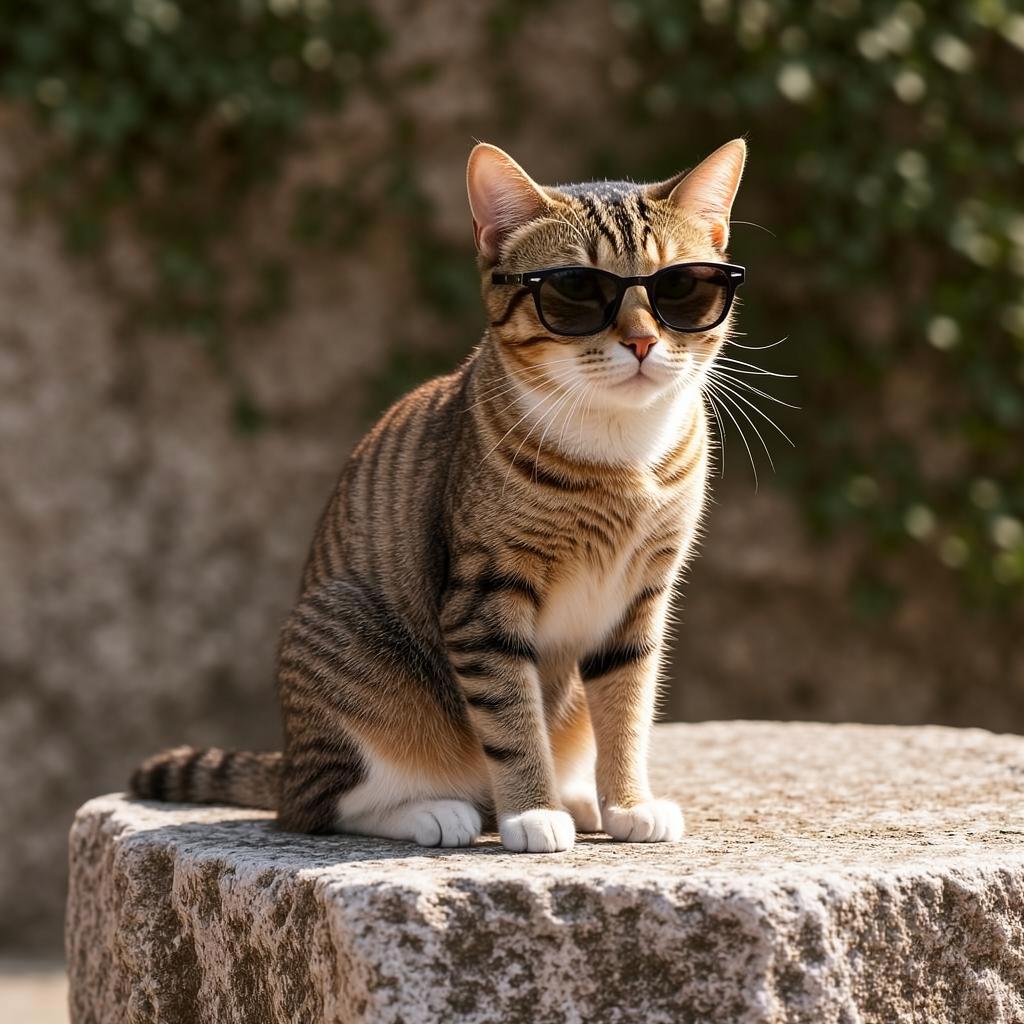}
        } \\
    \end{tabular}
    \caption{Local inpainting results. Prompt 1: ``An orange cat is sitting on a stone.'' Prompt 2: ``A cat wearing sunglasses is sitting on a stone.''}
    \label{fig:local_inpainting_outputs}
\end{figure}

Local inpainting is a specialized image-editing task in which the model receives both an input image and a mask specifying the region to be regenerated, with the objective of modifying only the masked area while preserving all remaining content. For this setting, we train a dedicated local-inpainting Template model. The model alone, however, provides only soft control and thus cannot strictly guarantee that unmasked regions remain completely unchanged, even though such failures are infrequent in practice. A key advantage of Diffusion Templates is that arbitrary pipeline inputs can be incorporated into the Template cache, making it possible to combine learned model-level control with pipeline-level hard constraints. Concretely, after each denoising step, we replace the unmasked region with the VAE encoding of the original image, thereby enforcing exact preservation of content outside the target area. As shown in Figure~\ref{fig:local_inpainting_outputs}, this simple pipeline-level constraint enables realistic local edits while maintaining stable and faithful reconstruction of the untouched regions.

\subsection{Age Control}
\begin{figure}[t]
    \centering
    \begin{tabular}{>{\centering\arraybackslash}m{0.31\linewidth} >{\centering\arraybackslash}m{0.31\linewidth} >{\centering\arraybackslash}m{0.31\linewidth}}
        Age: 20 & Age: 50 & Age: 80 \\
        \includegraphics[width=\linewidth]{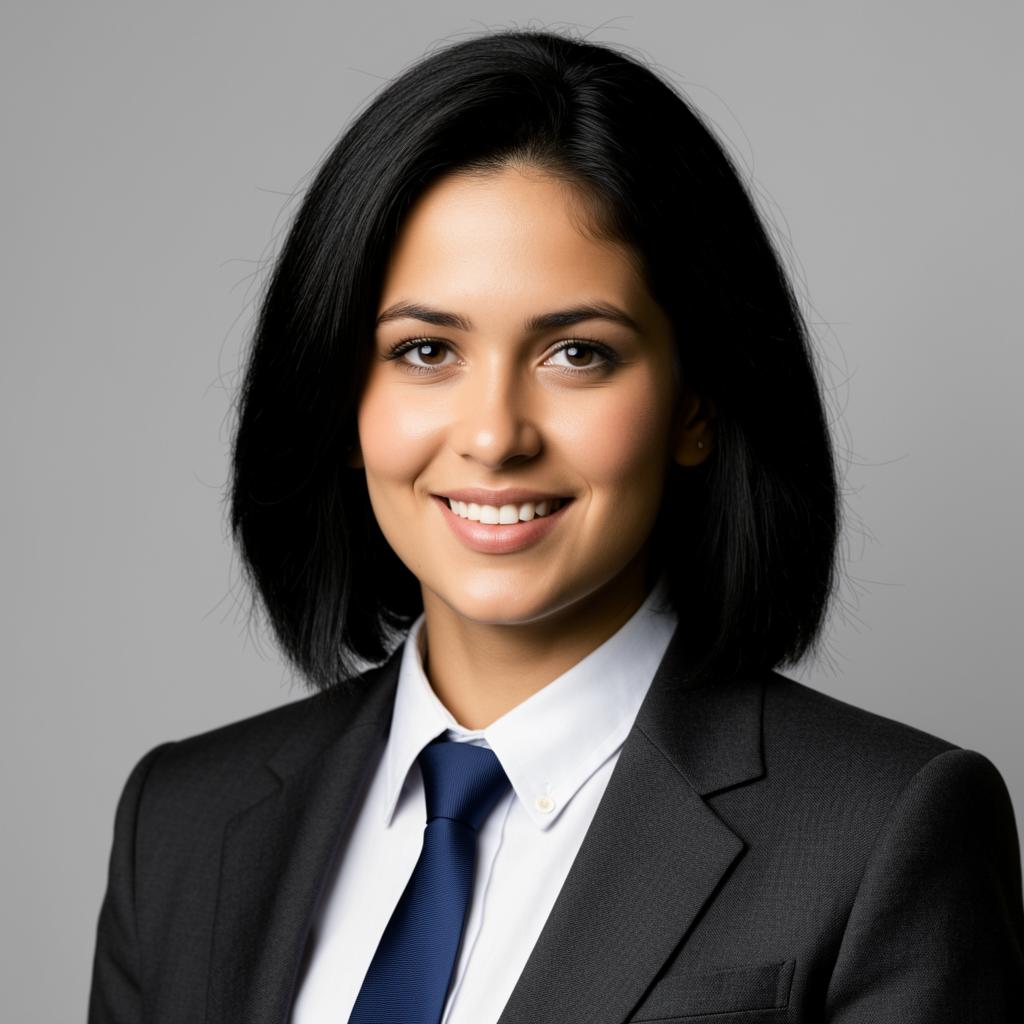} &
        \includegraphics[width=\linewidth]{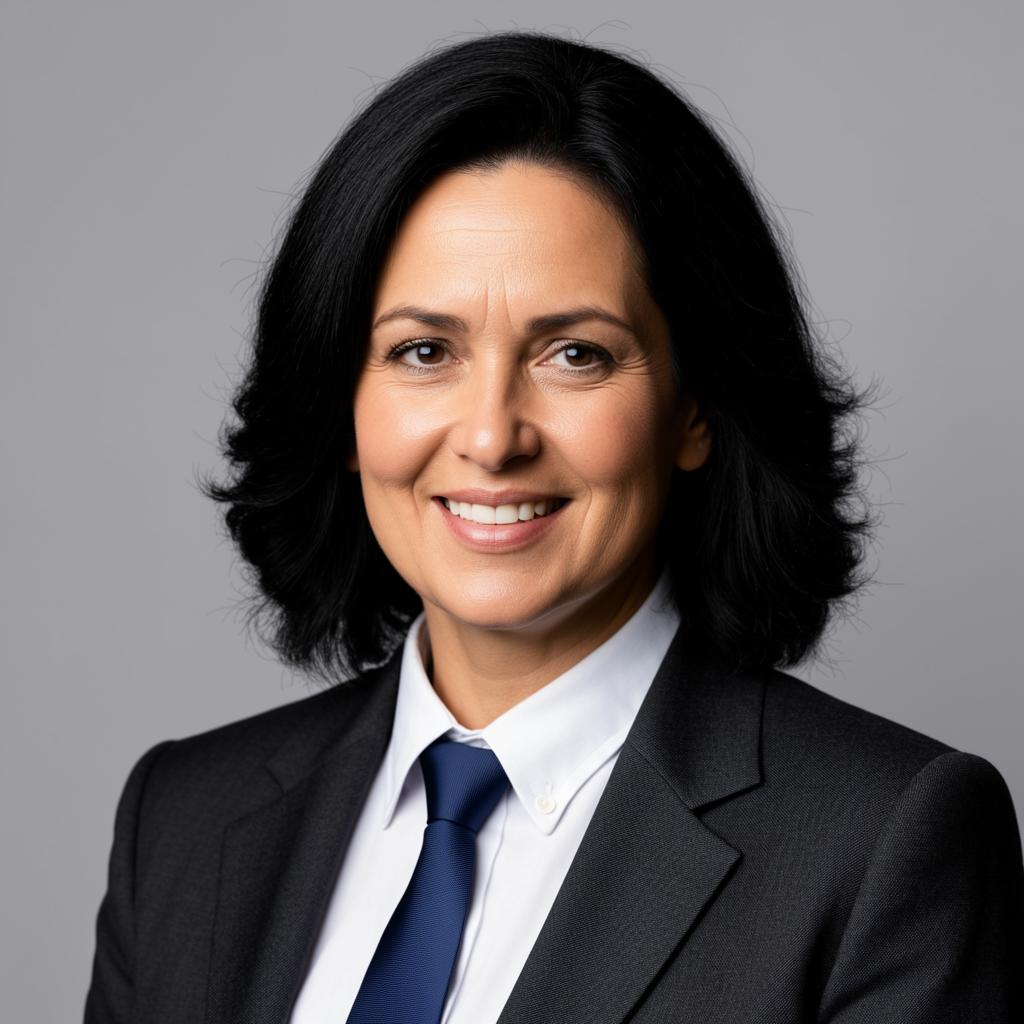} &
        \includegraphics[width=\linewidth]{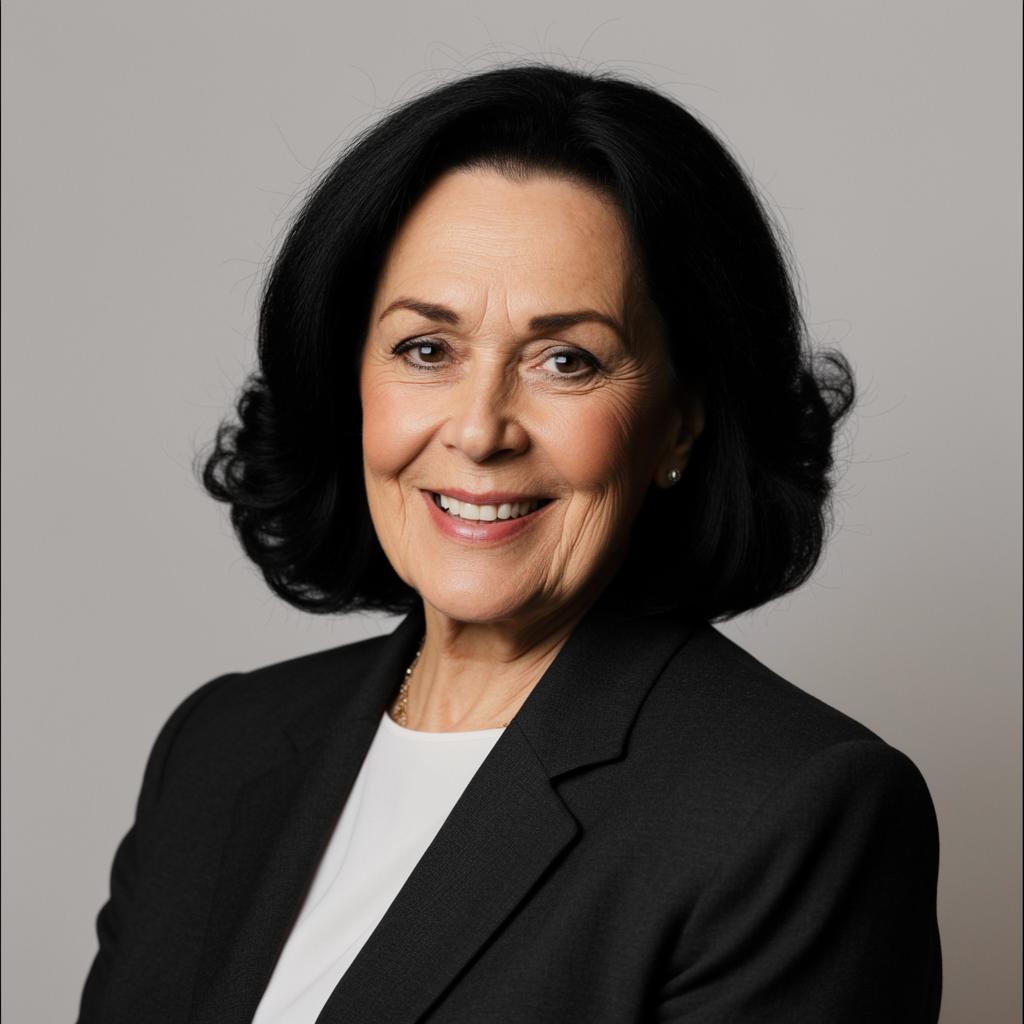} \\
    \end{tabular}
    \caption{Age control results with a shared prompt: ``A portrait of a woman with black hair, wearing a suit.''}
    \label{fig:age_control_table}
\end{figure}

We further evaluate the controllability of Template models in human portrait generation by training an age-control model on the IMDB-WIKI dataset~\cite{rothe2015dex}. This model adopts exactly the same architecture as the brightness-adjustment model, thereby providing a direct test of whether the same scalar-control formulation can be extended from low-level visual attributes to semantically richer human-specific attributes. The control signal is a scalar age value ranging from 10 to 90. Because the original dataset is imbalanced across age groups, we perform resampling over different age intervals to obtain a more balanced training distribution. As shown in Figure~\ref{fig:age_control_table}, the generated portraits exhibit a clear and coherent progression as the input age increases. In particular, age-related facial characteristics, such as wrinkles, become gradually more pronounced, while the overall identity and portrait quality remain stable. These results suggest that the proposed Template formulation is capable of learning meaningful and continuous control over age in portrait generation.

\subsection{Template Fusion}
\begin{figure}[htbp]
    \centering
    \includegraphics[width=\linewidth]{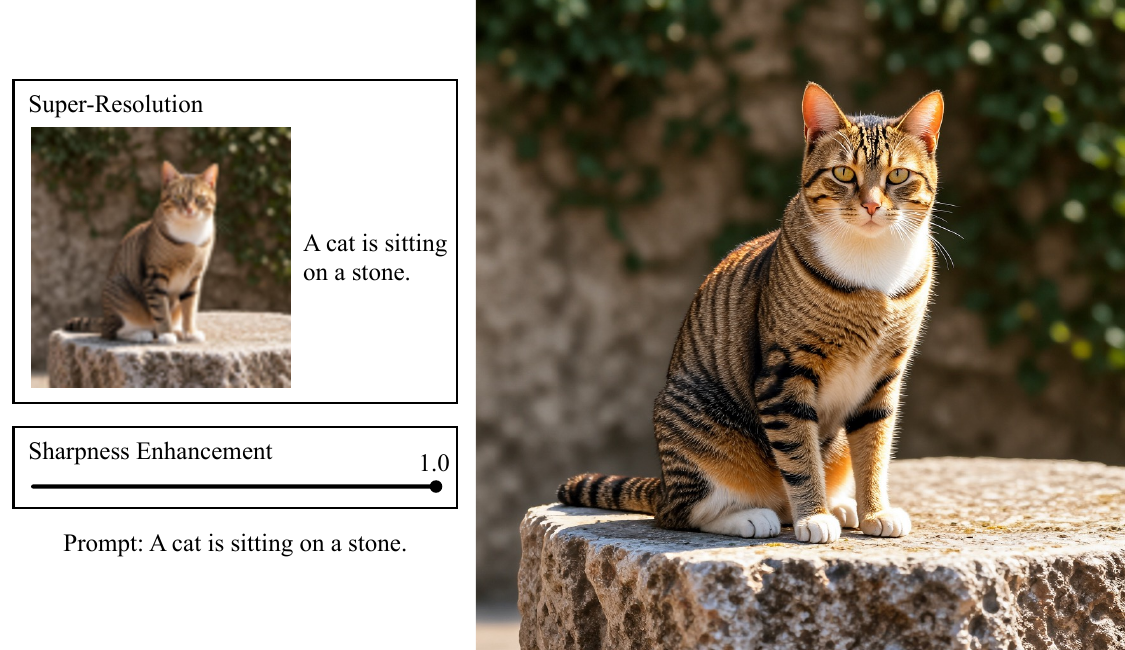}
    \caption{Fusion of super-resolution and sharpness enhancement capabilities, producing images with higher resolution and clearer details.}
    \label{fig:fusion_result_1}
\end{figure}

\begin{figure}[htbp]
    \centering
    \includegraphics[width=\linewidth]{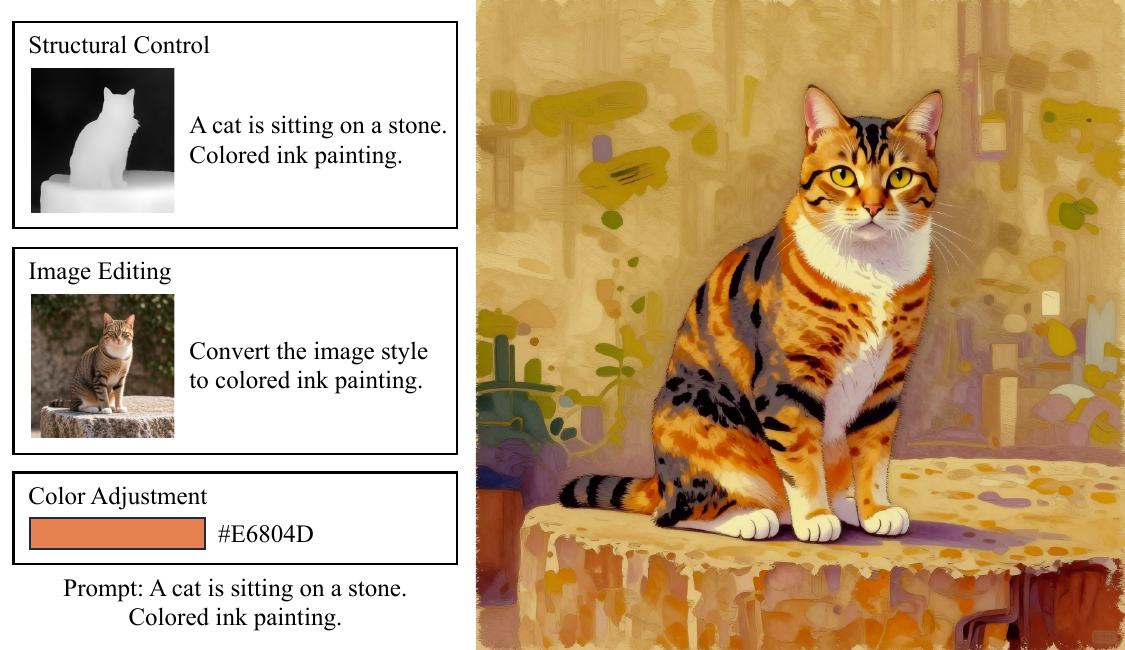}
    \caption{Fusion of structural control, image editing, and color adjustment, enabling the generation of artistic images with arbitrary tonal styles.}
    \label{fig:fusion_result_2}
\end{figure}

\begin{figure}[htbp]
    \centering
    \includegraphics[width=\linewidth]{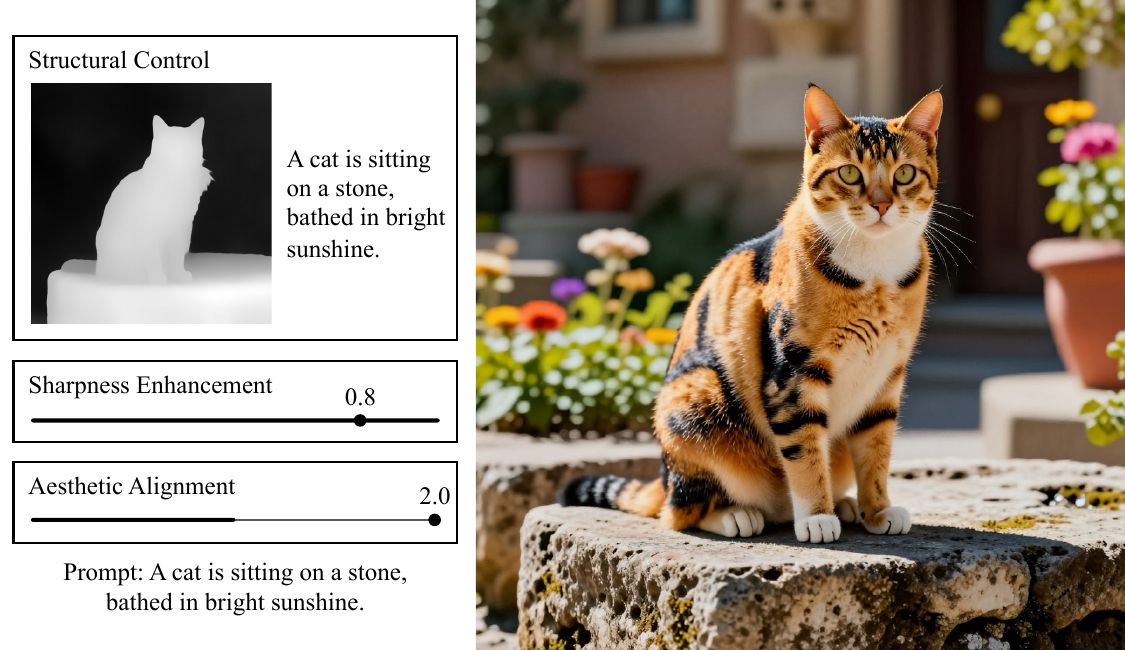}
    \caption{Fusion of structural control, sharpness enhancement, and aesthetic alignment, yielding renderings that better match human aesthetic preferences.}
    \label{fig:fusion_result_3}
\end{figure}

\begin{figure}[htbp]
    \centering
    \includegraphics[width=\linewidth]{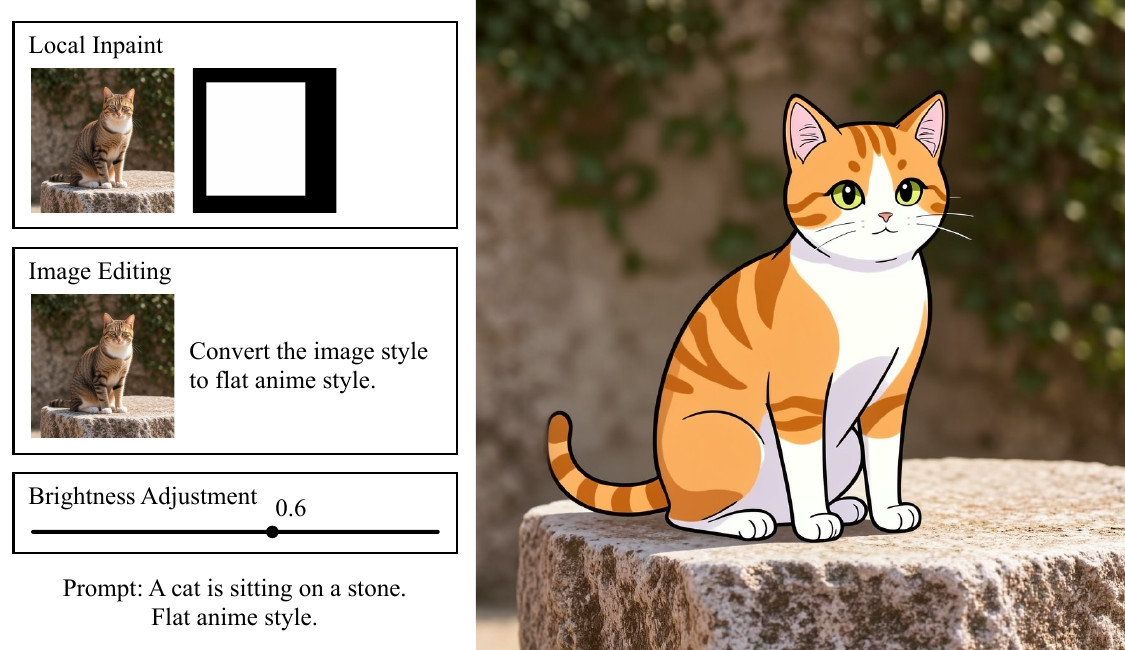}
    \caption{Fusion of local inpainting, image editing, and brightness adjustment, enabling localized changes to the visual style of the image.}
    \label{fig:fusion_result_4}
\end{figure}

Multiple Template models can be fused effectively within a single generation pipeline. The fusion strategy is determined by the format of the Template cache emitted by each model. For Template models that use KV-Cache as the cache representation, fusion can be implemented by concatenating caches along the sequence dimension. For models that use LoRA as the cache representation, fusion can be realized by concatenating the corresponding LoRA parameters along the rank dimension. When different Template models produce caches in heterogeneous formats, the associated modules can simply be enabled simultaneously, without requiring conversion to a unified representation. Moreover, because Template models themselves do not participate in the denoising loop of the diffusion model, the framework can leverage on-demand loading to support the fusion of an arbitrary number of model capabilities, without causing GPU memory consumption to grow substantially with the number of fused Template models. Figures~\ref{fig:fusion_result_1}, \ref{fig:fusion_result_2}, \ref{fig:fusion_result_3}, and \ref{fig:fusion_result_4} show several representative examples. These results suggest that Template fusion can yield more fine-grained and compositional control, thereby supporting a broader range of controllable generation scenarios.

\section{Conclusion and Future Work}

In this paper, we presented Diffusion Templates, a unified and open plugin framework for controllable diffusion models. By decoupling base-model inference from capability injection, the framework reformulates heterogeneous controllable generation methods as reusable Template models that communicate with the underlying diffusion runtime through a shared Template cache interface. This design improves modularity at both training and deployment time: new capabilities can be packaged independently, transferred across compatible backbones more easily, and composed within a common pipeline without repeatedly redesigning denoising internals. Across a diverse model zoo covering structural control, scalar attribute adjustment, image-conditioned editing, super-resolution, aesthetic alignment, content reference, local inpainting, and age control, our case studies demonstrate that the framework is flexible enough to unify a broad range of controllable generation tasks under one systems abstraction.

Diffusion Templates is still a prototype framework, and we plan to continue improving both its functionality and practical usability. Important directions for future work include:
\begin{itemize}
    \item Exploring efficient capability interfaces. Although KV-Cache and LoRA currently provide convenient and expressive interfaces for capability injection, other Template cache formats may offer better trade-offs in efficiency, compatibility, or controllability for different model architectures and downstream tasks.
    \item Extending the framework to a broader range of foundation models. In addition to supporting more image-generation backbones, we are particularly interested in adapting Diffusion Templates to video-generation models, where reusable capability interfaces may enable more flexible control over temporal consistency, motion patterns, and compositional structure.
    \item Evaluating these Template models quantitatively. While the current work mainly demonstrates the framework through representative qualitative examples, future studies should measure controllability, compositionality, transferability, efficiency, and compatibility under standardized benchmarks, so that the capabilities of different Template models can be compared more rigorously.
\end{itemize}

\bibliographystyle{plain}
\bibliography{refs}

@article{ho2020ddpm,
  title={Denoising diffusion probabilistic models},
  author={Ho, Jonathan and Jain, Ajay and Abbeel, Pieter},
  journal={Advances in neural information processing systems},
  volume={33},
  pages={6840--6851},
  year={2020}
}

@article{kingma2014vae,
  title={Auto-encoding variational bayes},
  author={Kingma, Diederik P and Welling, Max},
  journal={arXiv preprint arXiv:1312.6114},
  year={2013}
}

@inproceedings{ronneberger2015unet,
  title={U-net: Convolutional networks for biomedical image segmentation},
  author={Ronneberger, Olaf and Fischer, Philipp and Brox, Thomas},
  booktitle={International Conference on Medical image computing and computer-assisted intervention},
  pages={234--241},
  year={2015},
  organization={Springer}
}

@inproceedings{radford2021clip,
  title={Learning transferable visual models from natural language supervision},
  author={Radford, Alec and Kim, Jong Wook and Hallacy, Chris and Ramesh, Aditya and Goh, Gabriel and Agarwal, Sandhini and Sastry, Girish and Askell, Amanda and Mishkin, Pamela and Clark, Jack and others},
  booktitle={International conference on machine learning},
  pages={8748--8763},
  year={2021},
  organization={PmLR}
}

@article{raffel2020t5,
  title={Exploring the limits of transfer learning with a unified text-to-text transformer},
  author={Raffel, Colin and Shazeer, Noam and Roberts, Adam and Lee, Katherine and Narang, Sharan and Matena, Michael and Zhou, Yanqi and Li, Wei and Liu, Peter J},
  journal={Journal of machine learning research},
  volume={21},
  number={140},
  pages={1--67},
  year={2020}
}

@article{yang2025qwen3,
  title={Qwen3 technical report},
  author={Yang, An and Li, Anfeng and Yang, Baosong and Zhang, Beichen and Hui, Binyuan and Zheng, Bo and Yu, Bowen and Gao, Chang and Huang, Chengen and Lv, Chenxu and others},
  journal={arXiv preprint arXiv:2505.09388},
  year={2025}
}

@article{song2021ddim,
  title={Denoising diffusion implicit models},
  author={Song, Jiaming and Meng, Chenlin and Ermon, Stefano},
  journal={arXiv preprint arXiv:2010.02502},
  year={2020}
}

@article{ho2022cfg,
  title={Classifier-free diffusion guidance},
  author={Ho, Jonathan and Salimans, Tim},
  journal={arXiv preprint arXiv:2207.12598},
  year={2022}
}

@inproceedings{rombach2022ldm,
  title={High-resolution image synthesis with latent diffusion models},
  author={Rombach, Robin and Blattmann, Andreas and Lorenz, Dominik and Esser, Patrick and Ommer, Bj{\"o}rn},
  booktitle={Proceedings of the IEEE/CVF conference on computer vision and pattern recognition},
  pages={10684--10695},
  year={2022}
}

@article{podell2023sdxl,
  title={Sdxl: Improving latent diffusion models for high-resolution image synthesis},
  author={Podell, Dustin and English, Zion and Lacey, Kyle and Blattmann, Andreas and Dockhorn, Tim and M{\"u}ller, Jonas and Penna, Joe and Rombach, Robin},
  journal={arXiv preprint arXiv:2307.01952},
  year={2023}
}

@inproceedings{esser2024sd3,
  title={Scaling rectified flow transformers for high-resolution image synthesis},
  author={Esser, Patrick and Kulal, Sumith and Blattmann, Andreas and Entezari, Rahim and M{\"u}ller, Jonas and Saini, Harry and Levi, Yam and Lorenz, Dominik and Sauer, Axel and Boesel, Frederic and others},
  booktitle={Forty-first international conference on machine learning},
  year={2024}
}

@inproceedings{peebles2023dit,
  title={Scalable diffusion models with transformers},
  author={Peebles, William and Xie, Saining},
  booktitle={Proceedings of the IEEE/CVF international conference on computer vision},
  pages={4195--4205},
  year={2023}
}

@misc{blackforest2024flux,
  title={FLUX.1 Model Family},
  author={{Black Forest Labs}},
  year={2024},
  howpublished={Technical report/model release},
  note={\url{https://blackforestlabs.ai/}}
}

@article{wan2025wan,
  title={Wan: Open and advanced large-scale video generative models},
  author={Wan, Team and Wang, Ang and Ai, Baole and Wen, Bin and Mao, Chaojie and Xie, Chen-Wei and Chen, Di and Yu, Feiwu and Zhao, Haiming and Yang, Jianxiao and others},
  journal={arXiv preprint arXiv:2503.20314},
  year={2025}
}

@article{tencent2024hunyuanimage,
  title={Hunyuan-dit: A powerful multi-resolution diffusion transformer with fine-grained chinese understanding},
  author={Li, Zhimin and Zhang, Jianwei and Lin, Qin and Xiong, Jiangfeng and Long, Yanxin and Deng, Xinchi and Zhang, Yingfang and Liu, Xingchao and Huang, Minbin and Xiao, Zedong and others},
  journal={arXiv preprint arXiv:2405.08748},
  year={2024}
}

@article{chen2024pixart,
  title={Pixart-$\alpha$: Fast training of diffusion transformer for photorealistic text-to-image synthesis},
  author={Chen, Junsong and Yu, Jincheng and Ge, Chongjian and Yao, Lewei and Xie, Enze and Wu, Yue and Wang, Zhongdao and Kwok, James and Luo, Ping and Lu, Huchuan and others},
  journal={arXiv preprint arXiv:2310.00426},
  year={2023}
}

@article{xie2025sana,
  title={Sana: Efficient high-resolution image synthesis with linear diffusion transformers},
  author={Xie, Enze and Chen, Junsong and Chen, Junyu and Cai, Han and Tang, Haotian and Lin, Yujun and Zhang, Zhekai and Li, Muyang and Zhu, Ligeng and Lu, Yao and others},
  journal={arXiv preprint arXiv:2410.10629},
  year={2024}
}

@article{qwen2025qwenimage,
  title={Qwen-image technical report},
  author={Wu, Chenfei and Li, Jiahao and Zhou, Jingren and Lin, Junyang and Gao, Kaiyuan and Yan, Kun and Yin, Sheng-ming and Bai, Shuai and Xu, Xiao and Chen, Yilei and others},
  journal={arXiv preprint arXiv:2508.02324},
  year={2025}
}

@article{kong2025hunyuanvideo,
  title={Hunyuanvideo: A systematic framework for large video generative models},
  author={Kong, Weijie and Tian, Qi and Zhang, Zijian and Min, Rox and Dai, Zuozhuo and Zhou, Jin and Xiong, Jiangfeng and Li, Xin and Wu, Bo and Zhang, Jianwei and others},
  journal={arXiv preprint arXiv:2412.03603},
  year={2024}
}

@article{chen2025attrictrl,
  title={AttriCtrl: Fine-Grained Control of Aesthetic Attribute Intensity in Diffusion Models},
  author={Chen, Die and Duan, Zhongjie and Li, Zhiwen and Chen, Cen and Chen, Daoyuan and Li, Yaliang and Chen, Yingda},
  journal={arXiv preprint arXiv:2508.02151},
  year={2025}
}

@inproceedings{zhang2025eligen,
  title={Eligen: Entity-level controlled image generation with regional attention},
  author={Zhang, Hong and Duan, Zhongjie and Wang, Xingjun and Chen, Yingda and Zhang, Yu},
  booktitle={Proceedings of the 7th ACM International Conference on Multimedia in Asia},
  pages={1--7},
  year={2025}
}

@article{hu2022lora,
  title={Lora: Low-rank adaptation of large language models.},
  author={Hu, Edward J and Shen, Yelong and Wallis, Phillip and Allen-Zhu, Zeyuan and Li, Yuanzhi and Wang, Shean and Wang, Liang and Chen, Weizhu and others},
  journal={Iclr},
  volume={1},
  number={2},
  pages={3},
  year={2022}
}

@inproceedings{gal2022textualinversion,
  title={An Image is Worth One Word: Personalizing Text-to-Image Generation using Textual Inversion},
  author={Gal, Rinon and Alaluf, Yuval and Atzmon, Yuval and Patashnik, Or and Bermano, Amit Haim and Chechik, Gal and Cohen-Or, Daniel},
  booktitle={The Eleventh International Conference on Learning Representations},
  year={2023}
}

@inproceedings{ruiz2023dreambooth,
  title={Dreambooth: Fine tuning text-to-image diffusion models for subject-driven generation},
  author={Ruiz, Nataniel and Li, Yuanzhen and Jampani, Varun and Pritch, Yael and Rubinstein, Michael and Aberman, Kfir},
  booktitle={Proceedings of the IEEE/CVF conference on computer vision and pattern recognition},
  pages={22500--22510},
  year={2023}
}

@inproceedings{zhang2023controlnet,
  title={Adding conditional control to text-to-image diffusion models},
  author={Zhang, Lvmin and Rao, Anyi and Agrawala, Maneesh},
  booktitle={Proceedings of the IEEE/CVF international conference on computer vision},
  pages={3836--3847},
  year={2023}
}

@inproceedings{mou2023t2iadapter,
  title={T2i-adapter: Learning adapters to dig out more controllable ability for text-to-image diffusion models},
  author={Mou, Chong and Wang, Xintao and Xie, Liangbin and Wu, Yanze and Zhang, Jian and Qi, Zhongang and Shan, Ying},
  booktitle={Proceedings of the AAAI conference on artificial intelligence},
  volume={38},
  pages={4296--4304},
  year={2024}
}

@article{ye2023ipadapter,
  title={Ip-adapter: Text compatible image prompt adapter for text-to-image diffusion models},
  author={Ye, Hu and Zhang, Jun and Liu, Sibo and Han, Xiao and Yang, Wei},
  journal={arXiv preprint arXiv:2308.06721},
  year={2023}
}

@article{schick2023toolformer,
  title={Toolformer: Language models can teach themselves to use tools},
  author={Schick, Timo and Dwivedi-Yu, Jane and Dess{\`\i}, Roberto and Raileanu, Roberta and Lomeli, Maria and Hambro, Eric and Zettlemoyer, Luke and Cancedda, Nicola and Scialom, Thomas},
  journal={Advances in neural information processing systems},
  volume={36},
  pages={68539--68551},
  year={2023}
}

@inproceedings{yao2023react,
  title={React: Synergizing reasoning and acting in language models},
  author={Yao, Shunyu and Zhao, Jeffrey and Yu, Dian and Du, Nan and Shafran, Izhak and Narasimhan, Karthik R and Cao, Yuan},
  booktitle={The eleventh international conference on learning representations},
  year={2022}
}

@article{qin2023toollearning,
  title={Tool learning with foundation models},
  author={Qin, Yujia and Hu, Shengding and Lin, Yankai and Chen, Weize and Ding, Ning and Cui, Ganqu and Zeng, Zheni and Zhou, Xuanhe and Huang, Yufei and Xiao, Chaojun and others},
  journal={ACM Computing Surveys},
  volume={57},
  number={4},
  pages={1--40},
  year={2024},
  publisher={ACM New York, NY}
}

@article{xi2024agentsurvey,
  title={The rise and potential of large language model based agents: A survey},
  author={Xi, Zhiheng and Chen, Wenxiang and Guo, Xin and He, Wei and Ding, Yiwen and Hong, Boyang and Zhang, Ming and Wang, Junzhe and Jin, Senjie and Zhou, Enyu and others},
  journal={Science China Information Sciences},
  volume={68},
  number={2},
  pages={121101},
  year={2025},
  publisher={Springer}
}

@misc{openai2023functioncalling,
  title={Function Calling and Tool Use in OpenAI Models},
  author={{OpenAI}},
  year={2023},
  howpublished={Technical documentation},
  note={\url{https://platform.openai.com/docs/guides/function-calling}}
}

@misc{anthropic2024mcp,
  title={Model Context Protocol Specification},
  author={{Anthropic}},
  year={2024},
  howpublished={Technical specification},
  note={\url{https://modelcontextprotocol.io/}}
}

@misc{anthropic2025skills,
  title={Introducing Agent Skills},
  author={{Anthropic}},
  year={2025},
  howpublished={Product announcement},
  note={\url{https://www.anthropic.com/news/skills}, accessed April 12, 2026}
}

@article{dao2022flashattention,
  title={Flashattention: Fast and memory-efficient exact attention with io-awareness},
  author={Dao, Tri and Fu, Dan and Ermon, Stefano and Rudra, Atri and R{\'e}, Christopher},
  journal={Advances in neural information processing systems},
  volume={35},
  pages={16344--16359},
  year={2022}
}

@article{dao2023flashattention2,
  title={Flashattention-2: Faster attention with better parallelism and work partitioning},
  author={Dao, Tri},
  journal={arXiv preprint arXiv:2307.08691},
  year={2023}
}

@inproceedings{kwon2023vllm,
  title={Efficient memory management for large language model serving with pagedattention},
  author={Kwon, Woosuk and Li, Zhuohan and Zhuang, Siyuan and Sheng, Ying and Zheng, Lianmin and Yu, Cody Hao and Gonzalez, Joseph and Zhang, Hao and Stoica, Ion},
  booktitle={Proceedings of the 29th symposium on operating systems principles},
  pages={611--626},
  year={2023}
}

@article{zhang2023h2o,
  title={H2o: Heavy-hitter oracle for efficient generative inference of large language models},
  author={Zhang, Zhenyu and Sheng, Ying and Zhou, Tianyi and Chen, Tianlong and Zheng, Lianmin and Cai, Ruisi and Song, Zhao and Tian, Yuandong and R{\'e}, Christopher and Barrett, Clark and others},
  journal={Advances in Neural Information Processing Systems},
  volume={36},
  pages={34661--34710},
  year={2023}
}

@article{li2024snapkv,
  title={Snapkv: Llm knows what you are looking for before generation},
  author={Li, Yuhong and Huang, Yingbing and Yang, Bowen and Venkitesh, Bharat and Locatelli, Acyr and Ye, Hanchen and Cai, Tianle and Lewis, Patrick and Chen, Deming},
  journal={Advances in Neural Information Processing Systems},
  volume={37},
  pages={22947--22970},
  year={2024}
}

@article{srivatsa2024preble,
  title={Preble: Efficient distributed prompt scheduling for llm serving},
  author={Srivatsa, Vikranth and He, Zijian and Abhyankar, Reyna and Li, Dongming and Zhang, Yiying},
  journal={arXiv preprint arXiv:2407.00023},
  year={2024}
}

@article{abhyankar2024infercept,
  title={Infercept: Efficient intercept support for augmented large language model inference},
  author={Abhyankar, Reyna and He, Zijian and Srivatsa, Vikranth and Zhang, Hao and Zhang, Yiying},
  journal={arXiv preprint arXiv:2402.01869},
  year={2024}
}

@article{qin2024mooncake,
  title={Mooncake: A kvcache-centric disaggregated architecture for llm serving},
  author={Qin, Ruoyu and Li, Zheming and He, Weiran and Cui, Jialei and Tang, Heyi and Ren, Feng and Ma, Teng and Cai, Shangming and Zhang, Yineng and Zhang, Mingxing and others},
  journal={ACM Transactions on Storage},
  year={2024},
  publisher={ACM New York, NY}
}

@article{canny1986edge,
  title={A computational approach to edge detection},
  author={Canny, John},
  journal={IEEE Transactions on pattern analysis and machine intelligence},
  volume={8},
  number={6},
  pages={679--698},
  year={1986},
  publisher={Ieee}
}

@inproceedings{wang2021realesrgan,
  title={Real-esrgan: Training real-world blind super-resolution with pure synthetic data},
  author={Wang, Xintao and Xie, Liangbin and Dong, Chao and Shan, Ying},
  booktitle={Proceedings of the IEEE/CVF international conference on computer vision},
  pages={1905--1914},
  year={2021}
}

@article{jiang2024genaiarena,
  title={Genai arena: An open evaluation platform for generative models},
  author={Jiang, Dongfu and Ku, Max and Li, Tianle and Ni, Yuansheng and Sun, Shizhuo and Fan, Rongqi and Chen, Wenhu},
  journal={Advances in Neural Information Processing Systems},
  volume={37},
  pages={79889--79908},
  year={2024}
}

@article{kirstain2023pickapic,
  title={Pick-a-pic: An open dataset of user preferences for text-to-image generation},
  author={Kirstain, Yuval and Polyak, Adam and Singer, Uriel and Matiana, Shahbuland and Penna, Joe and Levy, Omer},
  journal={Advances in neural information processing systems},
  volume={36},
  pages={36652--36663},
  year={2023}
}

@article{duan2024artaug,
  title={ArtAug: Enhancing Text-to-Image Generation through Synthesis-Understanding Interaction},
  author={Duan, Zhongjie and Zhao, Qianyi and Chen, Cen and Chen, Daoyuan and Zhou, Wenmeng and Li, Yaliang and Chen, Yingda},
  journal={arXiv preprint arXiv:2412.12888},
  year={2024}
}

@article{tschannen2025siglip2,
  title={Siglip 2: Multilingual vision-language encoders with improved semantic understanding, localization, and dense features},
  author={Tschannen, Michael and Gritsenko, Alexey and Wang, Xiao and Naeem, Muhammad Ferjad and Alabdulmohsin, Ibrahim and Parthasarathy, Nikhil and Evans, Talfan and Beyer, Lucas and Xia, Ye and Mustafa, Basil and others},
  journal={arXiv preprint arXiv:2502.14786},
  year={2025}
}

@inproceedings{rothe2015dex,
  title={Dex: Deep expectation of apparent age from a single image},
  author={Rothe, Rasmus and Timofte, Radu and Van Gool, Luc},
  booktitle={Proceedings of the IEEE international conference on computer vision workshops},
  pages={10--15},
  year={2015}
}

@article{hacohen2026ltx2,
  title={LTX-2: Efficient Joint Audio-Visual Foundation Model},
  author={HaCohen, Yoav and Brazowski, Benny and Chiprut, Nisan and Bitterman, Yaki and Kvochko, Andrew and Berkowitz, Avishai and Shalem, Daniel and Lifschitz, Daphna and Moshe, Dudu and Porat, Eitan and others},
  journal={arXiv preprint arXiv:2601.03233},
  year={2026}
}

\end{document}